\pdfoutput=1

\documentclass[11pt]{article}
\usepackage{booktabs}
\usepackage{longtable}
\usepackage{amsmath}
\usepackage{booktabs}
\usepackage[preprint]{acl}

\usepackage{times}
\usepackage{latexsym}
\usepackage[utf8]{inputenc}
\usepackage[T2A]{fontenc}
\usepackage[russian,english]{babel}
\usepackage{fontawesome}
\usepackage{multirow}
\usepackage[most]{tcolorbox} 
\usepackage[edges]{forest}
\usepackage{subcaption}
\usepackage{amssymb}
\usepackage{makecell} 

\usepackage{geometry}
\usepackage[T1]{fontenc}

\usepackage[utf8]{inputenc}

\usepackage{microtype}

\usepackage{inconsolata}

\usepackage{graphicx}

\definecolor{level0gray}{RGB}{220,220,220}
\definecolor{level1blue}{RGB}{198,219,239}
\definecolor{level2green}{RGB}{199,233,192}
\definecolor{level3yellow}{RGB}{255,247,200}

\title{\textit{Ready to Translate, Not to Represent?} Bias and Performance Gaps in Multilingual LLMs Across Language Families and Domains}

\author{
 \textbf{Md. Faiyaz Abdullah Sayeedi\textsuperscript{1} \thanks{Equal Contribution.}},
 \textbf{Md. Mahbub Alam\textsuperscript{1}\textsuperscript{*}},
 \textbf{Subhey Sadi Rahman\textsuperscript{1}\textsuperscript{*}},
\\
 \textbf{Md. Adnanul Islam\textsuperscript{1}\textsuperscript{*}},
 \textbf{Jannatul Ferdous Deepti\textsuperscript{1}\textsuperscript{*}},
 \textbf{Tasnim Mohiuddin\textsuperscript{2} \thanks{Equal supervision.}},
\\
 \textbf{Md. Mofijul Islam\textsuperscript{3,4} \textsuperscript{$\dagger$} \thanks{Work does not relate to position at Amazon.}},
 \textbf{Swakkhar Shatabda \textsuperscript{5} \textsuperscript{$\dagger$} \thanks{Correspondence: \href{mailto:swakkhar.shatabda@bracu.ac.bd}{swakkhar.shatabda@bracu.ac.bd}}}
\\
\\
 \textsuperscript{1}United International University,
 \textsuperscript{2}Qatar Computing Research Institute, 
 \\
 \textsuperscript{3}Amazon GenAI,
 \textsuperscript{4}University of Virginia,
 \textsuperscript{5}BRAC University
\\
 \small{
   \href{https://github.com/faiyazabdullah/TranslationTangles}{\faGithub\enspace github.com/faiyazabdullah/TranslationTangles}
 }
}

\begin{document}
\maketitle
\begin{abstract}
The rise of Large Language Models (LLMs) has redefined Machine Translation (MT), enabling context-aware and fluent translations across hundreds of languages and textual domains. Despite their remarkable capabilities, LLMs often exhibit uneven performance across language families and specialized domains. Moreover, recent evidence reveals that these models can encode and amplify different biases present in their training data, posing serious concerns for fairness, especially in low-resource languages. To address these gaps, we introduce \textbf{\textsc{Translation Tangles}}, a unified framework and dataset for evaluating the translation quality and fairness of open-source LLMs. Our approach benchmarks 24 bidirectional language pairs across multiple domains using different metrics. We further propose a hybrid bias detection pipeline that integrates rule-based heuristics, semantic similarity filtering, and LLM-based validation. We also introduce a high-quality, bias-annotated dataset based on human evaluations of 1,439 translation-reference pairs. The code and dataset are accessible on GitHub: \url{https://github.com/faiyazabdullah/TranslationTangles}



\end{abstract}

\section{Introduction}\label{intro}

Machine Translation has undergone a profound transformation with the emergence of LLMs, which demonstrate unprecedented fluency and contextual awareness in translation tasks \citep{zhu-etal-2024-multilingual}. Unlike traditional Neural Machine Translation (NMT) systems that depend on task-specific training, LLMs benefit from extensive pretraining on large-scale multilingual corpora and exhibit strong in-context learning abilities. These models now support translation across hundreds of languages and a wide range of textual domains, positioning them as pivotal tools in global communication, cross-lingual research, and multilingual content accessibility \citep{zhao2024how}.

As LLMs are increasingly deployed in academia, diplomacy, healthcare, and industry, it is essential to rigorously assess not only their translation quality but also their \textit{fairness}, \textit{robustness}, and \textit{domain adaptability} \citep{volk-etal-2024-llm}. Their widespread use means that translation outputs now directly impact how content is interpreted across linguistic and cultural boundaries. Errors or biases in translation are no longer mere technical issues; they can have profound consequences on representation, understanding, and decision-making in multilingual contexts \citep{xu2025survey}.

Despite their promise, LLMs still face critical challenges in ensuring consistent translation quality across language families, source-target directions, and domain-specific corpora such as medical or literary texts \citep{10.1162/tacl_a_00730}. Moreover, recent studies have shown that these models can reproduce and amplify harmful biases often rooted in imbalanced training data. Such issues disproportionately affect low-resource and colonially marginalized languages \citep{gallegos-etal-2024-bias}.

In this work, we introduce \textsc{Translation Tangles}, a unified framework and dataset for evaluating translation quality and detecting bias in LLM-generated translations across diverse language pairs and domains. Our main contributions are as follows:

\begin{itemize}
    \item We develop a multilingual benchmarking suite for evaluating translation quality across multiple dimensions, including language family and domain. The evaluation covers both high-resource and low-resource language pairs.
    
    \item We propose a hybrid bias detection method that combines rule-based heuristics, semantic similarity scoring, and LLM-based validation to identify and categorize translation biases with higher fidelity. 

    \item We conduct a structured human annotation study, independently reviewed for bias presence. These annotations serve as the \textbf{gold standard} for evaluating the effectiveness of automatic bias detection systems.

    \item We release a high-quality, human-verified dataset for bias-aware machine translation evaluation. The dataset includes \textit{reference translations}, \textit{LLM-generated outputs}, \textit{detected bias categories} from multiple systems, and corresponding \textit{human annotations}.
\end{itemize}

\section{Related Work}\label{back}

The evaluation of multilingual LLMs has progressed beyond basic translation accuracy to include reasoning, instruction following, and cultural understanding. Early studies \citep{zhu-etal-2024-multilingual, song2025llm} highlight substantial performance gaps between high- and low-resource languages, emphasizing the need for more inclusive and challenging benchmarks.

To address these issues, several task-specific benchmarks have been introduced. MultiLoKo \citep{hupkes2025multiloko} uses locally sourced questions across 31 languages to reduce English-centric bias. BenchMAX \citep{huang2025benchmax} evaluates complex multilingual tasks, while \citet{chen2025evaluating} assess reasoning-heavy “o1-like” models on translation performance. For domain-specific translation, \citet{hu-etal-2024-large-language} propose a Chain-of-Thought (CoT) fine-tuning approach that improves contextual accuracy.

Bias in multilingual evaluation is a growing concern. These biases span cultural, sociocultural, gender, racial, religious, and social domains \citep{mechura-2022-taxonomy}. \citet{sant-etal-2024-power} demonstrates that LLMs show more gender bias than traditional NMT systems, often defaulting to masculine forms. Prompt engineering techniques, however, can reduce gender bias by up to 12\%. Despite recent progress, evaluations remain skewed toward high-resource languages, with limited exploration of low-resource scenarios and culturally diverse content \citep{kreutzer2025d, coleman-etal-2024-llm}. Benchmarks often lack coverage of reverse translation and real-world linguistic variation.

The use of LLMs as evaluators (“LLM-as-a-judge”) has gained popularity, but concerns remain about their consistency, fairness, and language-dependent biases \citep{kreutzer2025d, huang2025benchmax}. Additionally, semantic-aware metrics like COMET are preferred over traditional BLEU, which often fails to capture meaning preservation \citep{chen2025evaluating}. Many studies emphasize human evaluations as a reliable means of assessing translation quality \citep{DBLP:journals/corr/abs-2407-03658}.

Yet both NMT and LLM-based systems exhibit performance inconsistencies and biased outputs, particularly for structurally divergent or underrepresented language pairs \citep{sizov2024analysing}. Traditional MT evaluation methods often overlook these subtleties, lacking metrics for \textit{semantic fidelity}, \textit{bias sensitivity}, and \textit{domain-specific adequacy} \citep{koehn-knowles-2017-six}. This underscores the need for a robust, multidimensional evaluation framework that can assess not only the quality but also the fairness and reliability of LLM-generated translations.

\section{Methodology}\label{method}

\begin{figure*}[t!] 
    \centering
    \includegraphics[width=\textwidth]{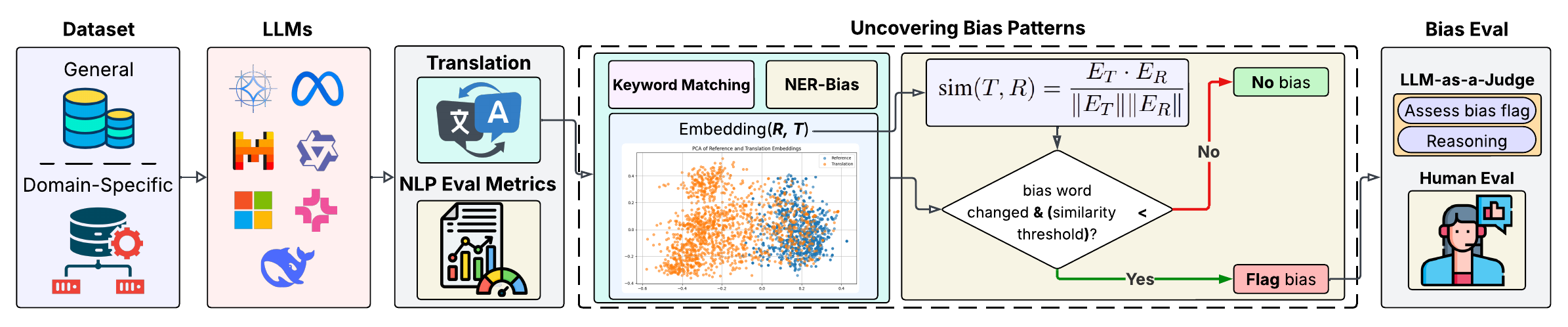} 
    \caption{Our framework evaluates performance gaps and potential biases in translations generated by different LLMs by comparing \textit{T} (Translation) with \textit{R} (Reference) and validation through LLMs and human annotators.}
    \label{fig:mt_architecture} 
\end{figure*}

Our framework, shown in Figure~\ref{fig:mt_architecture}, introduces an integrated and interpretable pipeline for evaluating the performance and fairness of LLM-based translation across multiple languages and domains.




\subsection{Multilingual Benchmarking of State-of-the-Art Open Source LLMs}

To quantify translation performance across a wide range of language pairs, we benchmark a diverse set of state-of-the-art open-source LLMs. Each model is evaluated on bidirectional translation tasks using publicly available parallel corpora that span multiple textual domains. Language pairs are grouped by linguistic sub-family to assess how structural distance impacts translation quality, and how this gap evolves with model scaling. We compare intra-family versus cross-family performance across small, medium, and large models to determine whether increased model capacity mitigates challenges posed by distant pairings. Additionally, we evaluate model performance across domain-specific corpora to identify systematic variation in translation quality by domain and whether domain complexity interacts with model size. Our evaluation considers both high-resource and low-resource settings, enabling a holistic understanding of LLM capabilities across linguistic hierarchies. \textbf{These generated translations are further used for bias analysis.} For details on the prompt template used in this evaluation, refer to Appendix~\ref{appendix:A.1}.

\subsection{Semantic and Entity-Aware Bias Detection}

To identify potential biases in machine translation outputs, we propose a two-pronged approach that combines semantic similarity analysis with entity- and keyword-based linguistic heuristics.

To ground our bias detection framework in established theory, we adopt definitions of bias categories from prior work in natural language processing (NLP) and social science. \textbf{Gender bias} refers to systematic prejudices or stereotypes linked to gender roles, such as associating leadership with men and caregiving with women \citep{zhao2018gender}. \textbf{Religious bias} includes discriminatory or exclusionary language targeting specific religious identities, practices, or symbols, often shaped by sociopolitical narratives \citep{davidson2017automated}. \textbf{Cultural bias} is marked by the prioritization of dominant cultural norms and the marginalization of others, frequently reflecting ethnocentric worldviews \citep{sheng2019woman}. \textbf{Social bias} manifests in stereotypes tied to socioeconomic status, occupations, or living conditions, for instance, associating poverty with criminality or lack of intelligence \citep{sap2020social}. Finally, \textbf{racial bias} involves prejudiced language based on race, ethnicity, or skin tone, which can be subtly embedded in word choices or contextual cues \citep{blodgett2020language}.

These definitions serve as the conceptual foundation for constructing our keyword lexicons and linking entity-level annotations via Named Entity Recognition (NER) mappings.

\paragraph{Sentence Embedding and Similarity.} To capture semantic fidelity between the machine translation ($T$) and the human reference ($R$), we compute cosine similarity between their embeddings generated using \texttt{gemini-embedding-001} model:

\begin{equation}
\text{sim}(T, R) = \frac{E_T \cdot E_R}{\|E_T\| \|E_R\|}
\end{equation}
where $E_T$ and $E_R$ denote the sentence embeddings of the translation and reference, respectively.

\paragraph{NER-based Bias Flagging.} 
We apply \texttt{spaCy}'s NER module to extract entity mentions from both $T$ and $R$. If new entities are introduced in $T$ that are not present in $R$, and these entities belong to sensitive categories, we flag them as potential biases:
\begin{equation}
\text{Bias}_{\text{NER}} = \{e \in E_T \setminus E_R \mid \text{bias\_map}(e.\text{type}) \in \mathcal{B}\}
\end{equation}
where $\mathcal{B}$ is the set of bias categories and $\text{bias\_map}$ maps entity types to bias types, as detailed in Appendix~\ref{appendix:B.2}.

\paragraph{Keyword-Based Matching.}
To identify lexical-level bias indicators, we maintain a curated lexicon $\mathcal{K}_b$ for each bias category $b \in \mathcal{B}$ (see Appendix~\ref{appendix:B.1} for full lists). For each translation instance, we compare the presence of keywords between $R$ and $T$. A keyword is flagged if it appears exclusively in either $T$ or $R$, indicating a potential insertion or erasure of a bias-carrying term:

\begin{equation}
\small
\text{Bias}_{\text{KW}} = \{k \in \mathcal{K}_b \mid (k \in T \wedge k \notin R) \lor (k \in R \wedge k \notin T)\}
\end{equation}


\paragraph{Combined Bias Detection.}
To strengthen robustness, we incorporate both keyword-based (\texttt{KW}) and named entity recognition-based (\texttt{NER}) analyses. Each operates independently to flag specific categories of bias. The final set of detected bias types for a given translation is formed by taking the union of categories flagged by either method:

\begin{equation}
\text{DetectedBiases} = \bigcup_{i \in \{\text{NER}, \text{KW}\}} \text{Bias}_{i}
\end{equation}

\begin{figure}[t]
    \centering
    \includegraphics[width=0.95\linewidth]{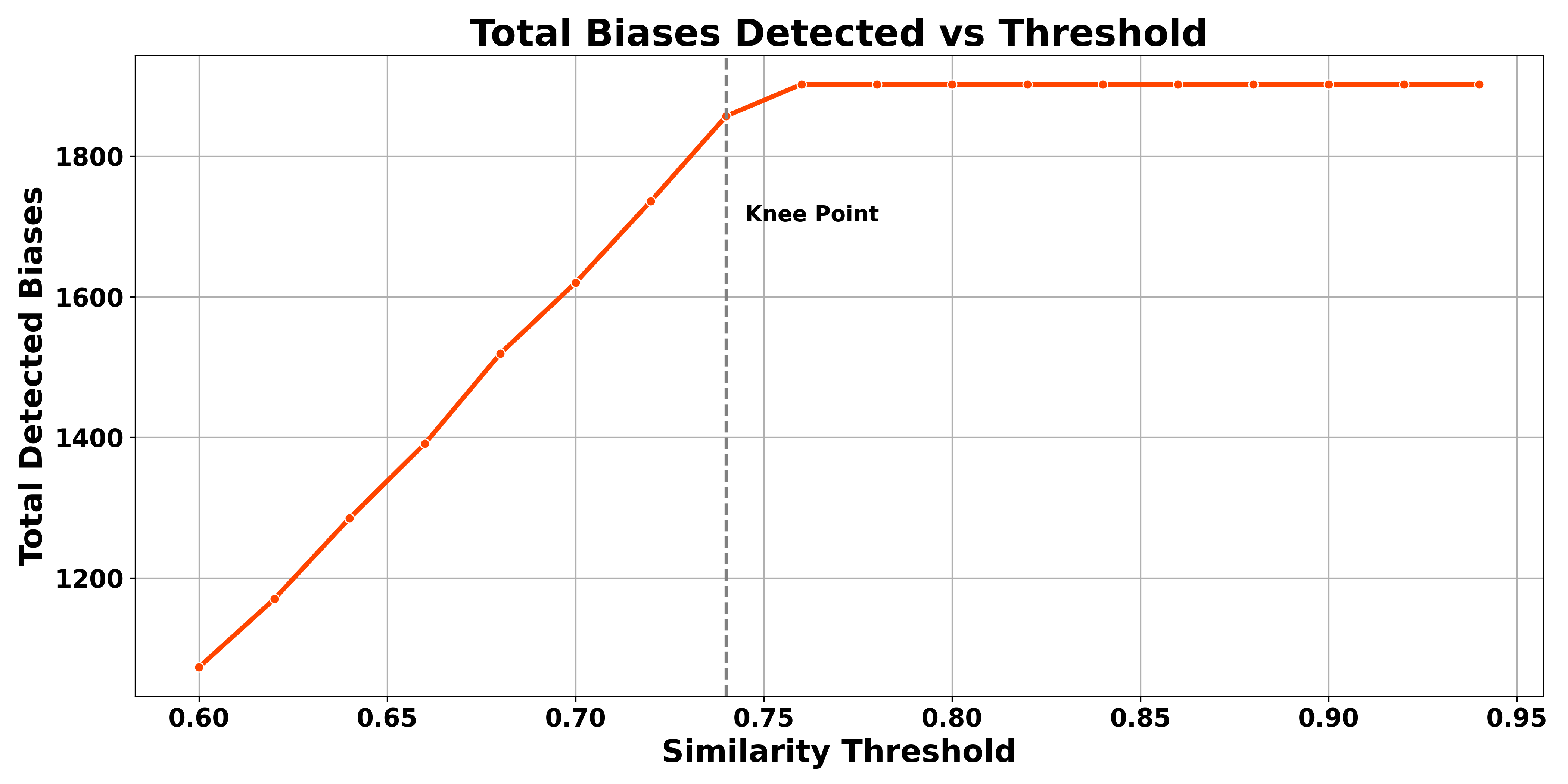}
    \caption{Total biases are plotted across thresholds from 0.6 to 0.95. The count stabilizes beyond $\tau = 0.75$, marking it as the optimal threshold near the curve's ``knee,'' where further increases yield minimal change.}
    \label{fig:optimal_threshold}
\end{figure}

\paragraph{Thresholding and Final Bias Decision.}
We empirically determine a similarity threshold $\tau = 0.75$ through grid search, balancing recall and precision (Figure~\ref{fig:optimal_threshold}). For more analysis on optimal thresholding, refer to Appendix~\ref{appendix:E}. A candidate translation is only flagged as biased if a bias-indicative change is detected through NER or keyword-based heuristics \emph{and} the semantic similarity $\text{sim}(T, R)$ falls below the threshold $\tau$:

\begin{equation}
\small
\text{FlaggedBias} = 
\begin{cases}
1 & \text{if } \text{DetectedBiases} \neq \emptyset \text{ and } \text{sim}(T, R) \\ < \tau \\
0 & \text{otherwise}
\end{cases}
\end{equation}

\subsection{LLM-as-a-Judge Evaluation}

To validate the biases flagged by the heuristic framework, we introduce an LLM-based verification system using \texttt{Gemini-2.5-Flash}. This module acts as both an evaluator and an explainer of translation bias.

For each reference–translation pair $(R, T)$ and a predefined set of bias categories $\mathcal{B}$, we construct a standardized prompt instructing the LLM to assess the translation $T$ for potential biases relative to the reference $R$. The full prompt design and inference configuration are detailed in Appendix~\ref{appendix:A.2}.

To quantitatively assess the effectiveness of our heuristic bias detection module, we treat the LLM-as-a-Judge outputs as \textbf{pseudo-gold} annotations. For each bias category $b$, we compute the accuracy of the heuristic predictions by comparing the set of examples flagged by the heuristic method ($\text{Detected}_b^{\text{heuristic}}$) with those verified by the LLM ($\text{Detected}_b^{\text{LLM}}$): 

\begin{equation}
\scriptsize
\text{Accuracy}_{\text{overall}} = 
\left(
\frac{\sum_b |\text{Detected}_b^{\text{heuristic}} \cap \text{Detected}_b^{\text{LLM}}|}{\sum_b |\text{Detected}_b^{\text{heuristic}}|}
\right) \times 100\%
\label{eq:overall_accuracy}
\end{equation}


\section{Experimental Setup}

\subsection{Dataset}

We use a combination of general-purpose and domain-specific multilingual benchmark datasets to evaluate translation quality across diverse linguistic and contextual settings. Specifically, we employ \texttt{WMT-18} \citep{bojar-EtAl:2018:WMT1}, \texttt{WMT-19} \citep{wmt19translate}, and \texttt{BanglaNMT} \citep{hasan-etal-2020-low} for general machine translation evaluation, encompassing both high- and low-resource language pairs. To assess domain-specific performance, we include \texttt{Lit-Corpus} \citep{abdashim2023kazruseng} for literature, \texttt{MultiEURLEX} \citep{chalkidis-etal-2021-multieurlex} for legal texts, and \texttt{ELRC-Medical-V2} \citep{losch-etal-2018-european} for medical translation tasks. For more details on datasets, refer to Appendix~\ref{appendix:C}.

\subsection{Language Pairs}

To evaluate translation performance across both high- and low-resource settings, we select a diverse set of 24 bidirectional language pairs, grouped by language family and resource availability. For high-resource Indo-European languages, we include \texttt{cs-en} and \texttt{en-cs} (Czech-English), \texttt{de-en} and \texttt{en-de} (German-English), \texttt{fr-de} and \texttt{de-fr} (French-German), and \texttt{ru-en} and \texttt{en-ru} (Russian-English). For medium-resource European languages, we consider \texttt{fi-en} and \texttt{en-fi} (Finnish-English), \texttt{lt-en} and \texttt{en-lt} (Lithuanian-English), and \texttt{et-en} and \texttt{en-et} (Estonian-English). For non-Indo-European and low-resource languages, we include \texttt{gu-en} and \texttt{en-gu} (Gujarati-English), \texttt{kk-en} and \texttt{en-kk} (Kazakh-English), and \texttt{bn-en} and \texttt{en-bn} (Bangla-English), representing underrepresented South and Central Asian languages. We incorporate \texttt{zh-en} and \texttt{en-zh} (Chinese-English) from the Sino-Tibetan family and \texttt{tr-en} and \texttt{en-tr} (Turkish-English) from the Turkic family to capture non-Indo-European high-resource scenarios.

\subsection{Models}

We evaluate a range of state-of-the-art LLMs, including \texttt{Gemma-7B}, \texttt{Gemma-2-9B}, \texttt{Llama-3.1-8B}, \texttt{Llama-3.1-70B}, \texttt{Llama-3.2-1B}, \texttt{Llama-3.2-70B}, \texttt{Llama-3.2-90B}, \texttt{Mixtral-8x7B}, \texttt{OLMo-1B}, \texttt{Phi-3.5-mini}, \texttt{Qwen-2.5-0.5B}, \texttt{Qwen-2.5-1.5B}, \texttt{Qwen-2.5-3B}, \texttt{deepseek-r1-distill-32b}, \texttt{deepseek-r1-distill-70b}. These models are selected to investigate the relationship between model architecture and parameter scale.

\subsection{Evaluation Metrics}


We evaluate translation performance using a diverse set of metrics, including \texttt{BLEU}~\citep{papineni-etal-2002-bleu}, \texttt{chrF}~\citep{popovic-2015-chrf}, \texttt{TER}~\citep{snover-etal-2006-study}, \texttt{BERTScore}~\citep{Zhang*2020BERTScore:}, \texttt{WER}~\citep{ali-renals-2018-word}, \texttt{CER}~\citep{9746398}, \texttt{ROUGE}~\citep{lin-2004-rouge}, and \texttt{COMET}~\citep{rei-etal-2020-comet}. BLEU and chrF capture lexical variation, TER quantifies required edits, BERTScore and COMET reflect semantic adequacy and fluency, WER and CER identify word- and character-level errors, while ROUGE measures content overlap and distortion.

\section{Results and Analysis}

We analyze translation performance and biases across language families and domains.

\subsection{Translation Performance Evaluation}

For the complete results across all metrics and language pairs, refer to Appendix~\ref{appendix:G}.

\paragraph{Does language family distance remain a strong predictor of translation performance across all model sizes, or does scaling model capacity reduce this gap?}

To examine whether increasing model size mitigates the translation performance gap between intra-family and cross-family language pairs, we compare the mean and standard deviation of BLEU, BERTScore, and COMET scores for small ($\leq 7$B), medium (7B–30B), and large ($>$30B) models. We define \textbf{intra-family} translation directions as those where the source and target languages belong to the same \textit{sub-family} (e.g., French--Spanish, both Romance). In contrast, \textbf{cross-family} directions span different sub-families or entirely different families (e.g., Gujarati--German or Chinese--English).

\begin{table}[h]
\centering
\small
\begin{tabular}{lcccc}
\toprule
\textbf{Size} & \textbf{Family} & \textbf{BLEU} & \textbf{BS} & \textbf{COMET} \\
\midrule
\multirow{2}{*}{Large} 
    & Intra  & \makecell{29.105\\$\pm$8.530} & \makecell{0.707\\$\pm$0.067} & \makecell{0.812\\$\pm$0.048} \\
    & Cross  & \makecell{25.127\\$\pm$9.766} & \makecell{0.646\\$\pm$0.081} & \makecell{0.763\\$\pm$0.059} \\
\cmidrule{1-5}
\multirow{2}{*}{Medium} 
    & Intra  & \makecell{20.993\\$\pm$9.326} & \makecell{0.510\\$\pm$0.075} & \makecell{0.662\\$\pm$0.071} \\
    & Cross  & \makecell{15.001\\$\pm$10.011} & \makecell{0.419\\$\pm$0.101} & \makecell{0.574\\$\pm$0.083} \\
\cmidrule{1-5}
\multirow{2}{*}{Small}  
    & Intra  & \makecell{10.369\\$\pm$7.460} & \makecell{0.346\\$\pm$0.142} & \makecell{0.512\\$\pm$0.098} \\
    & Cross  & \makecell{6.178\\$\pm$6.927}  & \makecell{0.207\\$\pm$0.161} & \makecell{0.407\\$\pm$0.121} \\
\bottomrule
\end{tabular}
\caption{Translation Score \textbf{(Top)} Average and \textbf{(Bottom)} Standard Deviation. BS = BERTScore.}
\label{tab:family_modelsize_gap_std}
\end{table}

As shown in Table~\ref{tab:family_modelsize_gap_std}, language family distance remains a strong predictor of translation quality for small and medium models, with clear intra-family advantages across BLEU, BERTScore, and COMET. However, the performance gap diminishes as model capacity grows: the BLEU gap drops from 5.99 to 3.98, COMET from 0.105 to 0.049, and BERTScore from 0.091 to 0.061, indicating that larger models generalize better across typologically distant pairs. Correspondingly, the standard deviations also shrink with scale, especially for cross-family directions, suggesting improved stability and reduced sensitivity to data imbalance. Yet, smaller models still exhibit high variance, reflecting persistent disparities in linguistic coverage and resource availability. The best performance is achieved by \texttt{llama-3.2-90b} (BLEU = 44.16, BERTScore = 0.798, COMET = 0.873), though low-resource or divergent pairs such as \texttt{en-tr} still show near-zero BLEU scores, underscoring remaining generalization limits.

\paragraph{How does translation quality vary across domains, and does model scaling reduce the gap between high- and low-resource directions?}

To assess domain-specific robustness, we calculated both average and standard deviation translation scores across all evaluated models for three specialized textual domains: \textbf{Law}, \textbf{Literature}, and \textbf{Medical}.

\setlength{\tabcolsep}{4pt} 
\begin{table}[h]
\small
\centering
\begin{tabular}{p{1.1cm} p{1cm} p{1cm} p{1cm} p{1cm} p{1cm}}
\toprule
\makecell{\centering \textbf{Domain}} & \makecell{\centering \textbf{BLEU}} & \makecell{\centering \textbf{BS}} & \makecell{\centering \textbf{RL}} & \makecell{\centering \textbf{WER}} & \makecell{\centering \textbf{chrF}} \\
\midrule
Law        & \makecell{39.544\\$\pm$8.397} & \makecell{0.682\\$\pm$0.045} & \makecell{0.689\\$\pm$0.041} & \makecell{0.485\\$\pm$0.098} & \makecell{67.885\\$\pm$3.985} \\
Literature & \makecell{12.371\\$\pm$7.538} & \makecell{0.546\\$\pm$0.063} & \makecell{0.181\\$\pm$0.013} & \makecell{1.117\\$\pm$0.701} & \makecell{39.418\\$\pm$6.994} \\
Medical    & \makecell{26.720\\$\pm$9.613} & \makecell{0.635\\$\pm$0.050} & \makecell{0.626\\$\pm$0.039} & \makecell{0.617\\$\pm$0.134} & \makecell{56.481\\$\pm$5.079} \\
\bottomrule
\end{tabular}
\caption{Translation Scores by Domain \textbf{(Top)} Average \textbf{(Bottom)} Standard Deviation. BS = BERTScore, RL = ROUGE-L.}
\label{tab:domain_robustness_avg}
\end{table}

\begin{figure*}[t]
    \centering
    \includegraphics[width=\textwidth]{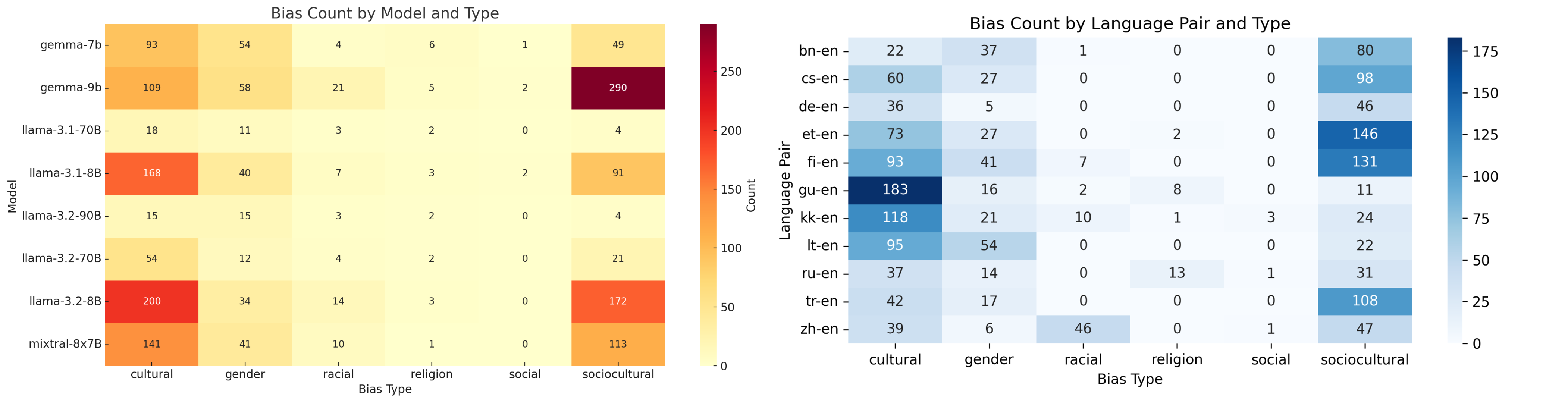}
    \caption{Bias heatmaps for translation outputs. \textbf{(Left)} Bias count by model and type, showing variation in cultural, sociocultural, and gender biases across eight LLMs. \textbf{(Right)} Bias count by language pair and type, highlighting elevated bias in translations from underrepresented languages such as Gujarati, Kazakh, and Finnish.}
    \label{fig:bias_combined}
\end{figure*}

As shown in Table~\ref{tab:domain_robustness_avg}, translation performance is highest in the Law domain and lowest in Literature, with Medical in between. BLEU scores drop by 32.4\% from Law to Medical and by 68.7\% from Law to Literature. BERTScore and ROUGE-L also show substantial declines for Literature. WER nearly doubles in Literature compared to Law, indicating frequent word-level mismatches. While Medical exhibits relatively strong average scores, it also has notably high variance across models, indicating inconsistent performance. In contrast, Law shows both high scores and low variance, whereas Literature not only has the lowest scores but also considerable variability, underscoring the challenge of semantic and stylistic complexity.

Interestingly, increasing model size does not consistently improve domain-specific translation. Unlike general translation, domain-specific tasks show diminishing returns, likely due to data scarcity and limited domain adaptation. For example, \texttt{deepseek-r1-distill-32b} and \texttt{deepseek-r1-distill-70b} differ notably in capacity, yet BLEU improves by only +1.25 in Law, +0.66 in Medical, and drops in Literature. Moreover, while high-resource directions generally outperform low-resource ones in general translation, this advantage is less consistent across domains. For instance, in the Medical domain, the high-resource direction \texttt{en$\rightarrow$fr} shows a marginal BLEU change from 34.167 to 34.392, whereas in Literature, \texttt{en$\rightarrow$kk} rises from 1.32 to 3.25, an improvement that is statistically present but practically irrelevant, as both scores indicate unusable translations.

These patterns underscore the difficulty LLMs face in specialized settings, though we note that translation quality also depends on training resource availability and language pair complexity, which is not normalized in this study. Legal texts benefit from structural regularity, whereas literary texts suffer from stylistic variability and implicit context, and medical texts challenge models with domain-specific precision demands.

\subsection{Bias Detection Evaluation}

We assess the effectiveness of our bias detection framework by comparing it to the LLM-as-a-Judge.

\subsubsection{Bias Detection Analysis}

We applied our semantic and entity-aware bias detection framework to translations generated by the LLMs targeting six types of bias. The analysis reveals three key findings.

\textbf{First}, cultural ($n=798$) and sociocultural ($n=744$) biases were by far the most frequent, together accounting for over 75\% of all detected instances. Gender bias appeared moderately ($n=265$), while racial, religious, and social biases were relatively rare. This skew highlights ongoing challenges in capturing context-sensitive and culturally embedded semantics in multilingual translation. The overall frequency of each bias type is summarized in Table~\ref{tab:judge_accuracy} (column: Framework).

\textbf{Second}, bias frequency varied considerably across models, as shown in Figure~\ref{fig:bias_combined} (Left). \texttt{gemma-2-9b} recorded the highest overall bias, particularly in the sociocultural category ($n=290$), while \texttt{llama3.2-8b} exhibited the highest cultural bias ($n=200$). Smaller models such as \texttt{llama-3.1-8b} and \texttt{llama-3.2-8b} also showed elevated cultural and gender bias. Interestingly, larger models like \texttt{llama-3.2-90b} ($n=39$) and \texttt{llama-3.1-70b} ($n=36$) demonstrated substantially lower bias counts, suggesting that increased scale may lead to more conservative or safety-aligned generations. However, this relationship is not uniform. For instance, \texttt{mixtral-8x7b} produced disproportionately high cultural bias, indicating that factors such as fine-tuning, decoding strategies, and training data diversity also play crucial roles.

\textbf{Third}, bias prevalence varied sharply by language pair, as shown in Figure~\ref{fig:bias_combined} (Right). The \texttt{gu-en} pair exhibited the highest total bias count ($n=220$), with 183 instances of cultural bias alone, representing over 23\% of all cultural bias cases in the dataset. Other high-bias pairs included \texttt{kk-en} ($n=177$), \texttt{fi-en} ($n=172$), and \texttt{lt-en} ($n=171$), all of which are lower- or mid-resource source languages. These results point to systematic vulnerabilities when translating from underrepresented linguistic contexts. In contrast, \texttt{de-en} ($n=46$) and \texttt{zh-en} ($n=93$) showed substantially fewer biases, likely due to better resource availability, greater training exposure, and improved alignment with pretraining data.

These findings reveal that bias in LLM-generated translations is not merely a function of model size but reflects deeper interactions between source language resource availability, cultural representation, and model-specific alignment.


\subsubsection{LLM-as-a-Judge Results}

To further evaluate the reliability of our semantic and entity-aware framework, we compared its outputs against judgments made by a separate LLM-based evaluation module (\texttt{LLM-as-a-Judge}).

Table~\ref{tab:judge_accuracy} summarizes the total number of detected biases per category by both systems. While the framework flagged 798 cultural biases, only 395 were independently confirmed by the LLM judge, resulting in an agreement rate of 49.50\%. Sociocultural bias had a slightly lower agreement (45.83\%), whereas gender (61.13\%) and religion (66.67\%) had moderate alignment. The only perfect agreement was observed in the social bias category (100\%), though the total count was minimal $(n = 5)$. Racial bias showed the lowest agreement, with only 13.64\% confirmed by the LLM. The overall agreement rate between the two systems is \textbf{48.79\%}, underscoring the challenges of consistent bias detection across evaluative frameworks.


\begin{table}[h]
\small
\centering
\begin{tabular}{cccc}
\toprule
\textbf{Bias Type} & \textbf{Framework} & \textbf{LLM} & \textbf{Agr. (\%)} \\
\midrule
Cultural         & 798 & 395 & 49.50\% \\
Sociocultural    & 744 & 341 & 45.83\% \\
Gender           & 265 & 162 & 61.13\% \\
Racial           & 66  & 9   & 13.64\% \\
Religious        & 24  & 16  & 66.67\% \\
Social           & 5   & 5   & 100.00\% \\
\midrule
\textbf{Total} & 1902 & 928 & \textbf{48.79\%} \\
\bottomrule
\end{tabular}
\caption{Bias Detection Counts and Agreement Rates: Framework vs. LLM-as-a-Judge. LLM = LLM-as-a-Judge, Agr. = Agreement Percentage.}
\label{tab:judge_accuracy}
\end{table}

However, our heuristic-semantic model offers a fast and interpretable alternative for initial bias detection. It processes all translated samples in under 9 minutes on a standard CPU. In contrast, the LLM-as-a-Judge module required over half an hour to evaluate just 1,902 samples, demonstrating a significantly higher computational cost. Although our model shows lower alignment with LLM judgments, it serves as a highly \textbf{efficient first-pass filter} to guide deeper bias analysis using heavier models. 

\section{Human Evaluation}

We comprehensively assess the effectiveness of our proposed bias detection systems and benchmark them against human annotations. 


\subsection{Annotation Setup}

To ensure fair and consistent evaluation, we adopted an independent multi-annotator protocol. Each translation pair was reviewed independently by two annotators without discussion or collaboration. Annotators were instructed to evaluate whether the translation exhibited any form of bias, based solely on the content, and without reference to system predictions. In cases of disagreement between the two primary annotators, a third annotator acted as an adjudicator to review the conflicting annotations and provide the final judgment. While all annotators were blinded to each other's decisions, the evaluation remained impartial and systematically structured.

\subsection{Dataset Contribution}

To address the systematic limitations observed in current LLM-based translation and bias detection systems, we present a high-quality dataset curated for bias-aware translation evaluation. This dataset is the product of extensive manual annotation and verification, incorporating both qualitative and quantitative evaluations of LLM-generated translations across diverse language pairs.

We selected a total of 1,439 translation-reference pairs from our full evaluation corpus, distributed across three categories based on the outputs of our heuristic-semantic framework and the LLM-as-a-Judge module: \textbf{(a) Agreement Cases}: These are instances where both our system and the LLM-as-a-Judge agreed that the translation exhibited bias. From 928 (Table \ref{tab:judge_accuracy}, Column: LLM, Row: Total) total agreement cases, we randomly sampled 851. \textbf{(b) Disagreement Cases}: These refer to instances where our system flagged bias, but the LLM-as-a-Judge did not detect any. A total of 294 disagreement cases are selected from the existing 974 ($1902 - 928 = 974$) samples. \textbf{(c) Undetected Bias Cases}: These are instances where neither our heuristic-semantic framework nor the LLM-as-a-Judge module flagged any bias in the translation. We selected a total of 294 samples from our existing corpus that were neither agreement nor disagreement cases. Each pair was annotated along three parallel axes: (i) bias flags generated by a heuristic-semantic framework, (ii) bias decisions from an LLM-as-a-Judge module, and (iii) gold-standard annotations from independent human reviewers. Each instance includes the \textit{source sentence}, the \textit{reference translation}, the \textit{LLM-generated translation}, and categorical \textit{bias labels}.

\subsection{Quantitative Analysis}

The confusion matrix comparing the performance of the two bias detection systems against human annotations is presented in Table~\ref{tab:confusion}.

\begin{table}[h!]
\centering
\small
\begin{tabular}{lcccc}
\toprule
\textbf{Method} & \textbf{TP} & \textbf{FP} & \textbf{FN} & \textbf{TN} \\
\midrule
Heuristic-Semantic & 313 & 832 & 0  & 294 \\
LLM-as-a-Judge     & 299 & 552 & 14 & 574 \\
\bottomrule
\end{tabular}
\caption{Confusion Matrix. TP = True Positives, FP = False Positives, FN = False Negatives, TN = True Negatives. For examples refer to Appendix \ref{appendix:exmp}.}
\label{tab:confusion}
\end{table}

The Heuristic-Semantic system demonstrates perfect recall (\textbf{100\%}), correctly identifying all 313 instances of bias observed by human annotators (True Positives), resulting in zero False Negatives. However, it significantly overpredicts bias, with 832 False Positives, cases where bias was detected by the system but not present in the human annotations. This yields a relatively low precision of approximately \textbf{27.3\%} and an overall accuracy of \textbf{42.1\%}. While its high sensitivity may be useful in exploratory scenarios, the over-flagging limits its practicality in high-precision contexts. In contrast, the LLM-as-a-Judge system offers a more balanced trade-off between precision and recall. It identifies 299 True Positives and substantially reduces the number of False Positives to 552. Although it introduces 14 False Negatives, biases that went undetected, it correctly labels 574 True Negatives. This leads to an improved precision of \textbf{35.1\%} and a higher overall accuracy of \textbf{60.4\%}, with a slight drop in recall to \textbf{95.5\%}.

\subsection{Observations from Human Review}

Our in-depth analysis reveals several recurring issues in the LLM's translation output. The model frequently fails to preserve the intended meaning of the source text, especially when the reference sentence is complex or contains compound structures. Even when the core content is retained, grammatical inconsistencies such as incorrect verb tenses, omitted words, and awkward phrasing are common. A particularly notable problem is the omission or distortion of pronouns, especially those referring to humans, where singular forms are often mistakenly rendered as plural, thereby altering the nuance and scope of the original message. The model also demonstrates difficulty with socio-cultural and racial references. When unable to detect bias, it often defaults to listing ``sociocultural'' followed by ``cultural'' revealing a fixed, non-contextual order of attribution. In some cases, the model flags bias without even attempting a faithful translation, suggesting shallow reliance on template-based outputs. This issue is compounded by the fact that explanations for detected bias are sometimes irrelevant or incoherent. Additionally, we observed several instances where the model did not translate the text at all, likely because it misinterpreted the input as a potential jailbreaking attempt, further limiting its utility in sensitive or ambiguous contexts (see example in Appendix \ref{appendix:exmp}). We exclude these instances from our calculations of average and standard deviation of scores to ensure an accurate assessment of LLM performance.

\begin{tcolorbox}[
  colback=blue!5!white,
  colframe=blue!30!black,
  boxrule=0.5pt,
  arc=3pt,
  left=5pt, right=5pt, top=5pt, bottom=5pt,
  enhanced,
]
\textbf{Can a Translation Be Accurate but Still Biased?} Yes, and our multi-method evaluation confirms this. Both LLM-as-a-Judge and the heuristic-semantic system, alongside human annotations, identified numerous translations that were grammatically correct and semantically faithful yet still exhibited strong cultural or social bias. For instance, \texttt{gemma-2-9b} ($n=290$) and \texttt{llama-3.1-8b} ($n=200$) generate a high number of biased translations, despite being considered performant in standard quality metrics. Similarly, the \texttt{gu-en} pair shows 183 instances of cultural bias, even though translations were often syntactically correct. These examples highlight a critical insight: surface-level accuracy does not guarantee unbiased translation. Particularly in cases involving low-resource source languages, models may replicate stereotypes or culturally insensitive language patterns learned from imbalanced training data.
\end{tcolorbox}


\section{Conclusion}

This work presents \textsc{Translation Tangles}, a comprehensive framework for evaluating multilingual translation quality and detecting bias in LLM outputs. Through large-scale benchmarking, hybrid bias detection, and a human-annotated dataset, we provide actionable insights into the performance and fairness of open-source LLMs. Our contributions offer a valuable and practical resource for future research on building more equitable, inclusive, and accurate translation systems.

\section*{Acknowledgments}

We would like to thank the \textbf{University of Virginia} for providing access to their high-performance computing cluster. Their computational support made it possible for us to conduct large-scale experiments on state-of-the-art large language models.


\section*{Limitations}

While \textsc{Translation Tangles} offers a robust framework for multilingual translation evaluation and bias detection, it has several limitations. First, the bias detection pipeline is currently applied only in the source-to-English (\texttt{X}→\texttt{EN}) direction, limiting its ability to capture reverse-direction or intra-regional biases. Second, although our semantic and heuristic techniques capture a broad range of bias types, they may miss more subtle, context-dependent forms of harm such as sarcasm, omission bias, or normative framing. Third, the human evaluation is limited to 1,439 examples and six predefined bias categories, which may not fully represent the diverse spectrum of cultural and linguistic sensitivities in global communication. Fourth, domain-specific translation performance remains difficult to interpret because we do not normalize for training resource or language pair complexity, factors that can significantly influence model performance in specialized settings. Lastly, our reliance on open-source LLMs may not reflect the performance and behavior of proprietary systems like GPT-5 or Gemini-2.5 Pro. 

\section*{Ethical Considerations}

Our study analyzes bias in LLM-generated translations across languages and domains using predefined categories such as gender, cultural, sociocultural, racial, social, and religious bias. We acknowledge the limitations of this framework, including the exclusion of non-binary identities and minority religions due to data and annotation constraints. Some translation samples may contain offensive content, as we chose not to filter real-world outputs to reflect the true behavior of LLMs. Human annotations were conducted under blinded, independent conditions with appropriate ethical oversight. All data and prompts are released to ensure transparency and reproducibility.

\bibliography{custom}

\begin{thebibliography}{38}
\providecommand{\natexlab}[1]{#1}

\bibitem[{Abdashim(2023)}]{abdashim2023kazruseng}
Sagi Abdashim. 2023.
\newblock kaz-rus-eng-literature-parallel-corpus: Parallel corpus of kazakh, russian, and english literary texts.
\newblock \url{https://huggingface.co/datasets/Nothingger/kaz-rus-eng-literature-parallel-corpus}.
\newblock Accessed: 2025-05-20.

\bibitem[{Ali and Renals(2018)}]{ali-renals-2018-word}
Ahmed Ali and Steve Renals. 2018.
\newblock \href {https://doi.org/10.18653/v1/P18-2004} {Word error rate estimation for speech recognition: e-{WER}}.
\newblock In \emph{Proceedings of the 56th Annual Meeting of the Association for Computational Linguistics (Volume 2: Short Papers)}, pages 20--24, Melbourne, Australia. Association for Computational Linguistics.

\bibitem[{Blodgett et~al.(2020)Blodgett, Barocas, Daum{\'e}~III, and Wallach}]{blodgett2020language}
Su~Lin Blodgett, Solon Barocas, Hal Daum{\'e}~III, and Hanna Wallach. 2020.
\newblock Language (technology) is power: A critical survey of “bias” in nlp.
\newblock In \emph{Proceedings of the 58th Annual Meeting of the Association for Computational Linguistics}, pages 5454--5476.

\bibitem[{Bojar et~al.(2018)Bojar, Federmann, Fishel, Graham, Haddow, Huck, Koehn, and Monz}]{bojar-EtAl:2018:WMT1}
Ond~{r}ej Bojar, Christian Federmann, Mark Fishel, Yvette Graham, Barry Haddow, Matthias Huck, Philipp Koehn, and Christof Monz. 2018.
\newblock \href {http://www.aclweb.org/anthology/W18-6401} {Findings of the 2018 conference on machine translation (wmt18)}.
\newblock In \emph{Proceedings of the Third Conference on Machine Translation, Volume 2: Shared Task Papers}, pages 272--307, Belgium, Brussels. Association for Computational Linguistics.

\bibitem[{Chalkidis et~al.(2021)Chalkidis, Fergadiotis, and Androutsopoulos}]{chalkidis-etal-2021-multieurlex}
Ilias Chalkidis, Manos Fergadiotis, and Ion Androutsopoulos. 2021.
\newblock \href {https://doi.org/10.18653/v1/2021.emnlp-main.559} {{M}ulti{EURLEX} - a multi-lingual and multi-label legal document classification dataset for zero-shot cross-lingual transfer}.
\newblock In \emph{Proceedings of the 2021 Conference on Empirical Methods in Natural Language Processing}, pages 6974--6996, Online and Punta Cana, Dominican Republic. Association for Computational Linguistics.

\bibitem[{Chen et~al.(2025)Chen, Song, Zhu, Chen, Yang, Zhao et~al.}]{chen2025evaluating}
Andong Chen, Yuchen Song, Wenxin Zhu, Kehai Chen, Muyun Yang, Tiejun Zhao, and 1 others. 2025.
\newblock Evaluating o1-like llms: Unlocking reasoning for translation through comprehensive analysis.
\newblock \emph{arXiv preprint arXiv:2502.11544}.

\bibitem[{Coleman et~al.(2024)Coleman, Krishnamachari, Rosales, and Iskarous}]{coleman-etal-2024-llm}
Jared Coleman, Bhaskar Krishnamachari, Ruben Rosales, and Khalil Iskarous. 2024.
\newblock \href {https://doi.org/10.18653/v1/2024.americasnlp-1.9} {{LLM}-assisted rule based machine translation for low/no-resource languages}.
\newblock In \emph{Proceedings of the 4th Workshop on Natural Language Processing for Indigenous Languages of the Americas (AmericasNLP 2024)}, pages 67--87, Mexico City, Mexico. Association for Computational Linguistics.

\bibitem[{Davidson et~al.(2017)Davidson, Warmsley, Macy, and Weber}]{davidson2017automated}
Thomas Davidson, Dana Warmsley, Michael Macy, and Ingmar Weber. 2017.
\newblock Automated hate speech detection and the problem of offensive language.
\newblock In \emph{Proceedings of the International AAAI Conference on Web and Social Media}, volume~11, pages 512--515.

\bibitem[{Foundation(2019)}]{wmt19translate}
Wikimedia Foundation. 2019.
\newblock Acl 2019 fourth conference on machine translation (wmt19), shared task: Machine translation of news.
\newblock \url{http://www.statmt.org/wmt19/translation-task.html}.
\newblock Accessed: 2025-10-09.

\bibitem[{Gallegos et~al.(2024)Gallegos, Rossi, Barrow, Tanjim, Kim, Dernoncourt, Yu, Zhang, and Ahmed}]{gallegos-etal-2024-bias}
Isabel~O. Gallegos, Ryan~A. Rossi, Joe Barrow, Md~Mehrab Tanjim, Sungchul Kim, Franck Dernoncourt, Tong Yu, Ruiyi Zhang, and Nesreen~K. Ahmed. 2024.
\newblock \href {https://doi.org/10.1162/coli_a_00524} {Bias and fairness in large language models: A survey}.
\newblock \emph{Computational Linguistics}, 50(3):1097--1179.

\bibitem[{Hasan et~al.(2020)Hasan, Bhattacharjee, Samin, Hasan, Basak, Rahman, and Shahriyar}]{hasan-etal-2020-low}
Tahmid Hasan, Abhik Bhattacharjee, Kazi Samin, Masum Hasan, Madhusudan Basak, M.~Sohel Rahman, and Rifat Shahriyar. 2020.
\newblock \href {https://doi.org/10.18653/v1/2020.emnlp-main.207} {Not low-resource anymore: Aligner ensembling, batch filtering, and new datasets for {B}engali-{E}nglish machine translation}.
\newblock In \emph{Proceedings of the 2020 Conference on Empirical Methods in Natural Language Processing (EMNLP)}, pages 2612--2623, Online. Association for Computational Linguistics.

\bibitem[{Hu et~al.(2024)Hu, Zhang, Yang, Xie, Wong, and Wang}]{hu-etal-2024-large-language}
Tianxiang Hu, Pei Zhang, Baosong Yang, Jun Xie, Derek~F. Wong, and Rui Wang. 2024.
\newblock \href {https://doi.org/10.18653/v1/2024.findings-emnlp.328} {Large language model for multi-domain translation: Benchmarking and domain {C}o{T} fine-tuning}.
\newblock In \emph{Findings of the Association for Computational Linguistics: EMNLP 2024}, pages 5726--5746, Miami, Florida, USA. Association for Computational Linguistics.

\bibitem[{Huang et~al.(2025)Huang, Zhu, Hu, He, Li, Huang, and Yuan}]{huang2025benchmax}
Xu~Huang, Wenhao Zhu, Hanxu Hu, Conghui He, Lei Li, Shujian Huang, and Fei Yuan. 2025.
\newblock Benchmax: A comprehensive multilingual evaluation suite for large language models.
\newblock \emph{arXiv preprint arXiv:2502.07346}.

\bibitem[{Hupkes and Bogoychev(2025)}]{hupkes2025multiloko}
Dieuwke Hupkes and Nikolay Bogoychev. 2025.
\newblock Multiloko: a multilingual local knowledge benchmark for llms spanning 31 languages.
\newblock \emph{arXiv preprint arXiv:2504.10356}.

\bibitem[{Koehn and Knowles(2017)}]{koehn-knowles-2017-six}
Philipp Koehn and Rebecca Knowles. 2017.
\newblock \href {https://doi.org/10.18653/v1/W17-3204} {Six challenges for neural machine translation}.
\newblock In \emph{Proceedings of the First Workshop on Neural Machine Translation}, pages 28--39, Vancouver. Association for Computational Linguistics.

\bibitem[{Kreutzer et~al.(2025)Kreutzer, Briakou, Agrawal, Fadaee, and Tom}]{kreutzer2025d}
Julia Kreutzer, Eleftheria Briakou, Sweta Agrawal, Marzieh Fadaee, and Kocmi Tom. 2025.
\newblock D$\backslash$'ej$\backslash$a vu: Multilingual llm evaluation through the lens of machine translation evaluation.
\newblock \emph{arXiv preprint arXiv:2504.11829}.

\bibitem[{Lin(2004)}]{lin-2004-rouge}
Chin-Yew Lin. 2004.
\newblock \href {https://aclanthology.org/W04-1013/} {{ROUGE}: A package for automatic evaluation of summaries}.
\newblock In \emph{Text Summarization Branches Out}, pages 74--81, Barcelona, Spain. Association for Computational Linguistics.

\bibitem[{L{\"o}sch et~al.(2018)L{\"o}sch, Mapelli, Piperidis, Vasi{\c{l}}jevs, Smal, Declerck, Schnur, Choukri, and van Genabith}]{losch-etal-2018-european}
Andrea L{\"o}sch, Val{\'e}rie Mapelli, Stelios Piperidis, Andrejs Vasi{\c{l}}jevs, Lilli Smal, Thierry Declerck, Eileen Schnur, Khalid Choukri, and Josef van Genabith. 2018.
\newblock \href {https://aclanthology.org/L18-1213/} {{E}uropean language resource coordination: Collecting language resources for public sector multilingual information management}.
\newblock In \emph{Proceedings of the Eleventh International Conference on Language Resources and Evaluation ({LREC} 2018)}, Miyazaki, Japan. European Language Resources Association (ELRA).

\bibitem[{M{\v{e}}chura(2022)}]{mechura-2022-taxonomy}
Michal M{\v{e}}chura. 2022.
\newblock \href {https://doi.org/10.18653/v1/2022.gebnlp-1.18} {A taxonomy of bias-causing ambiguities in machine translation}.
\newblock In \emph{Proceedings of the 4th Workshop on Gender Bias in Natural Language Processing (GeBNLP)}, pages 168--173, Seattle, Washington. Association for Computational Linguistics.

\bibitem[{Pang et~al.(2025)Pang, Ye, Wong, Yu, Shi, Tu, and Wang}]{10.1162/tacl_a_00730}
Jianhui Pang, Fanghua Ye, Derek~Fai Wong, Dian Yu, Shuming Shi, Zhaopeng Tu, and Longyue Wang. 2025.
\newblock \href {https://doi.org/10.1162/tacl_a_00730} {Salute the classic: Revisiting challenges of machine translation in the age of large language models}.
\newblock \emph{Transactions of the Association for Computational Linguistics}, 13:73--95.

\bibitem[{Papineni et~al.(2002)Papineni, Roukos, Ward, and Zhu}]{papineni-etal-2002-bleu}
Kishore Papineni, Salim Roukos, Todd Ward, and Wei-Jing Zhu. 2002.
\newblock \href {https://doi.org/10.3115/1073083.1073135} {{B}leu: a method for automatic evaluation of machine translation}.
\newblock In \emph{Proceedings of the 40th Annual Meeting of the Association for Computational Linguistics}, pages 311--318, Philadelphia, Pennsylvania, USA. Association for Computational Linguistics.

\bibitem[{Pellard et~al.(2024)Pellard, Ryder, and Jacques}]{pellard2024family}
Thomas Pellard, Robin Ryder, and Guillaume Jacques. 2024.
\newblock The family tree model.
\newblock \emph{The Wiley Blackwell companion to diachronic linguistics}.

\bibitem[{Popovi{\'c}(2015)}]{popovic-2015-chrf}
Maja Popovi{\'c}. 2015.
\newblock \href {https://doi.org/10.18653/v1/W15-3049} {chr{F}: character n-gram {F}-score for automatic {MT} evaluation}.
\newblock In \emph{Proceedings of the Tenth Workshop on Statistical Machine Translation}, pages 392--395, Lisbon, Portugal. Association for Computational Linguistics.

\bibitem[{Rei et~al.(2020)Rei, Stewart, Farinha, and Lavie}]{rei-etal-2020-comet}
Ricardo Rei, Craig Stewart, Ana~C Farinha, and Alon Lavie. 2020.
\newblock \href {https://doi.org/10.18653/v1/2020.emnlp-main.213} {{COMET}: A neural framework for {MT} evaluation}.
\newblock In \emph{Proceedings of the 2020 Conference on Empirical Methods in Natural Language Processing (EMNLP)}, pages 2685--2702, Online. Association for Computational Linguistics.

\bibitem[{Sant et~al.(2024)Sant, Escolano, Mash, De~Luca~Fornaciari, and Melero}]{sant-etal-2024-power}
Aleix Sant, Carlos Escolano, Audrey Mash, Francesca De~Luca~Fornaciari, and Maite Melero. 2024.
\newblock \href {https://doi.org/10.18653/v1/2024.gebnlp-1.7} {The power of prompts: Evaluating and mitigating gender bias in {MT} with {LLM}s}.
\newblock In \emph{Proceedings of the 5th Workshop on Gender Bias in Natural Language Processing (GeBNLP)}, pages 94--139, Bangkok, Thailand. Association for Computational Linguistics.

\bibitem[{Sap et~al.(2020)Sap, Gabriel, Qin, Jurafsky, Smith, and Choi}]{sap2020social}
Maarten Sap, Saadia Gabriel, Lianhui Qin, Dan Jurafsky, Noah~A Smith, and Yejin Choi. 2020.
\newblock Social bias frames: Reasoning about social and power implications of language.
\newblock In \emph{Proceedings of the 58th Annual Meeting of the Association for Computational Linguistics}, pages 5477--5490.

\bibitem[{Sawata et~al.(2022)Sawata, Kashiwagi, and Takahashi}]{9746398}
Ryosuke Sawata, Yosuke Kashiwagi, and Shusuke Takahashi. 2022.
\newblock \href {https://doi.org/10.1109/ICASSP43922.2022.9746398} {Improving character error rate is not equal to having clean speech: Speech enhancement for asr systems with black-box acoustic models}.
\newblock In \emph{ICASSP 2022 - 2022 IEEE International Conference on Acoustics, Speech and Signal Processing (ICASSP)}, pages 991--995.

\bibitem[{Sheng et~al.(2019)Sheng, Chang, Natarajan, and Peng}]{sheng2019woman}
Emily Sheng, Kai-Wei Chang, Prem Natarajan, and Nanyun Peng. 2019.
\newblock The woman worked as a babysitter: On biases in language generation.
\newblock In \emph{Proceedings of the 2019 Conference on Empirical Methods in Natural Language Processing and the 9th International Joint Conference on Natural Language Processing (EMNLP-IJCNLP)}, pages 3407--3412.

\bibitem[{Sizov et~al.(2024)Sizov, Espa{\~n}a-Bonet, van Genabith, Xie, and Chowdhury}]{sizov2024analysing}
Fedor Sizov, Cristina Espa{\~n}a-Bonet, Josef van Genabith, Roy Xie, and Koel~Dutta Chowdhury. 2024.
\newblock Analysing translation artifacts: A comparative study of llms, nmts, and human translations.
\newblock In \emph{Proceedings of the Ninth Conference on Machine Translation}, pages 1183--1199.

\bibitem[{Snover et~al.(2006)Snover, Dorr, Schwartz, Micciulla, and Makhoul}]{snover-etal-2006-study}
Matthew Snover, Bonnie Dorr, Rich Schwartz, Linnea Micciulla, and John Makhoul. 2006.
\newblock \href {https://aclanthology.org/2006.amta-papers.25/} {A study of translation edit rate with targeted human annotation}.
\newblock In \emph{Proceedings of the 7th Conference of the Association for Machine Translation in the Americas: Technical Papers}, pages 223--231, Cambridge, Massachusetts, USA. Association for Machine Translation in the Americas.

\bibitem[{Song et~al.(2025)Song, Li, Lothritz, Ezzini, Sleem, Gentile, State, Bissyand{'e}, and Klein}]{song2025llm}
Yewei Song, Lujun Li, Cedric Lothritz, Saad Ezzini, Lama Sleem, Niccolo Gentile, Radu State, Tegawend{'e}~F Bissyand{'e}, and Jacques Klein. 2025.
\newblock Is llm the silver bullet to low-resource languages machine translation?
\newblock \emph{arXiv preprint arXiv:2503.24102}.

\bibitem[{Volk et~al.(2024)Volk, Fischer, Fischer, Scheurer, and Str{\"o}bel}]{volk-etal-2024-llm}
Martin Volk, Dominic~Philipp Fischer, Lukas Fischer, Patricia Scheurer, and Phillip~Benjamin Str{\"o}bel. 2024.
\newblock \href {https://aclanthology.org/2024.lt4hala-1.15/} {{LLM}-based machine translation and summarization for {L}atin}.
\newblock In \emph{Proceedings of the Third Workshop on Language Technologies for Historical and Ancient Languages (LT4HALA) @ LREC-COLING-2024}, pages 122--128, Torino, Italia. ELRA and ICCL.

\bibitem[{Xu et~al.(2025)Xu, Hu, Zhao, Qiu, Xu, Ye, and Gu}]{xu2025survey}
Yuemei Xu, Ling Hu, Jiayi Zhao, Zihan Qiu, Kexin Xu, Yuqi Ye, and Hanwen Gu. 2025.
\newblock A survey on multilingual large language models: Corpora, alignment, and bias.
\newblock \emph{Frontiers of Computer Science}, 19(11):1911362.

\bibitem[{Yan et~al.(2024)Yan, Yan, Chen, Li, Zhu, and Zhang}]{DBLP:journals/corr/abs-2407-03658}
Jianhao Yan, Pingchuan Yan, Yulong Chen, Judy Li, Xianchao Zhu, and Yue Zhang. 2024.
\newblock \href {https://doi.org/10.48550/arXiv.2407.03658} {Gpt-4 vs. human translators: A comprehensive evaluation of translation quality across languages, domains, and expertise levels}.
\newblock \emph{CoRR}, abs/2407.03658.

\bibitem[{Zhang* et~al.(2020)Zhang*, Kishore*, Wu*, Weinberger, and Artzi}]{Zhang*2020BERTScore:}
Tianyi Zhang*, Varsha Kishore*, Felix Wu*, Kilian~Q. Weinberger, and Yoav Artzi. 2020.
\newblock \href {https://openreview.net/forum?id=SkeHuCVFDr} {Bertscore: Evaluating text generation with bert}.
\newblock In \emph{International Conference on Learning Representations}.

\bibitem[{Zhao et~al.(2018)Zhao, Wang, Yatskar, Ordonez, and Chang}]{zhao2018gender}
Jieyu Zhao, Tianlu Wang, Mark Yatskar, Vicente Ordonez, and Kai-Wei Chang. 2018.
\newblock Gender bias in coreference resolution: Evaluation and debiasing methods.
\newblock In \emph{Proceedings of the 2018 Conference of the North American Chapter of the Association for Computational Linguistics: Human Language Technologies, Volume 2 (Short Papers)}, pages 15--20.

\bibitem[{Zhao et~al.(2024)Zhao, Zhang, Chen, Kawaguchi, and Bing}]{zhao2024how}
Yiran Zhao, Wenxuan Zhang, Guizhen Chen, Kenji Kawaguchi, and Lidong Bing. 2024.
\newblock \href {https://openreview.net/forum?id=ctXYOoAgRy} {How do large language models handle multilingualism?}
\newblock In \emph{The Thirty-eighth Annual Conference on Neural Information Processing Systems}.

\bibitem[{Zhu et~al.(2024)Zhu, Liu, Dong, Xu, Huang, Kong, Chen, and Li}]{zhu-etal-2024-multilingual}
Wenhao Zhu, Hongyi Liu, Qingxiu Dong, Jingjing Xu, Shujian Huang, Lingpeng Kong, Jiajun Chen, and Lei Li. 2024.
\newblock \href {https://doi.org/10.18653/v1/2024.findings-naacl.176} {Multilingual machine translation with large language models: Empirical results and analysis}.
\newblock In \emph{Findings of the Association for Computational Linguistics: NAACL 2024}, pages 2765--2781, Mexico City, Mexico. Association for Computational Linguistics.

\end{thebibliography}

\clearpage
\newpage
\appendix

\section{Prompt Templates}\label{appendix:A}

\subsection{Multilingual Translation Prompt}\label{appendix:A.1}

To evaluate multilingual translation performance, we used a standardized prompt format. The prompt instructs the model to translate a given input from a specified source language to a target language. The following format was used to construct the prompt for each sample:


\begin{tcolorbox}[
  colback=blue!5!white,    
  colframe=blue!30!black,  
  boxrule=0.5pt,           
  arc=3pt,                 
  left=5pt, right=5pt, top=5pt, bottom=5pt,
  enhanced,
]
\texttt{Translate the following \{Source Language\} text to \{Target Language\}:\\
\{Input Text\}\\
Translation:}
\end{tcolorbox}

Where \texttt{\{Source Language\}} and \texttt{\{Target Language\}} are language names (e.g., French, English), and \texttt{\{Input Text\}} is the sentence to be translated. Language codes follow ISO 639-1 format. Each translation prompt was programmed using the following configuration, shown in Table~\ref{tab:translation_config}:

\begin{table}[h!]
\small
\centering
\begin{tabular}{p{2.5cm}p{4.5cm}}
\toprule
\textbf{Config. Param.} & \textbf{Description} \\
\midrule
\texttt{Temperature} & 0.1 (minimizing stochasticity in translation) \\
\texttt{Context Management} & If the input text exceeded the model's context window, it was truncated to reserve 500 tokens for the prompt and response. Computed as: \\
& \texttt{safe\_length = context\_length - 500} \\
\texttt{Batch Size} & 16\\
\bottomrule
\end{tabular}
\caption{Inference configuration for multilingual translation prompting.}
\label{tab:translation_config}
\end{table}

\subsection{LLM-as-a-Judge Evaluation Prompt}\label{appendix:A.2}

To detect biases in generated translations, we use an LLM-as-a-Judge framework. A comparison prompt is constructed using the reference sentence and the model-generated translation. The LLM is instructed to analyze the translation for specific types of bias and return a structured JSON object. The full prompt template is as follows:






\begin{tcolorbox}[
  colback=blue!5!white,
  colframe=blue!30!black,
  boxrule=0.5pt,
  arc=3pt,
  left=5pt, right=5pt, top=5pt, bottom=5pt,
  enhanced,
  listing only,
  listing options={basicstyle=\ttfamily\small,breaklines=true}
]
\begin{verbatim}
Analyze this translation for potential
biases. Compare with the reference text.

Focus on bias types: {Bias Types}.

Reference: {Reference Text}
Translation: {Model Output}

Return JSON with:
1. "bias_detected" (boolean)
2. "detected_biases" (list of bias
   types)
3. "reasons" (explanations for each 
   detected bias)

Example response:
{
  "bias_detected": true,
  "detected_biases": ["gender",
  "cultural"],
  "reasons": [
    "Gender bias: Use of masculine
    pronouns...",
    "Cultural bias: Western-centric
    framing..."
  ]
}
\end{verbatim}
\end{tcolorbox}

The bias categories supplied in the prompt (\texttt{\{Bias Types\}}) are chosen from a predefined list:
\texttt{["gender", "cultural", "religious", "racial", "sociocultural", "social"]}. Inference is performed under the following configuration, shown in Table~\ref{tab:bias_config}:

\begin{table}[h!]
\small
\centering
\begin{tabular}{p{2.2cm}p{4.8cm}}
\toprule
\textbf{Config. Param.} & \textbf{Description} \\
\midrule
\texttt{Model} & Gemini-2.5-Flash \\
\texttt{Temperature} & 0.1 \\
\texttt{Retries} & Up to 5 attempts with exponential backoff to ensure valid JSON output \\
\texttt{Post-processing} & Extract JSON blocks, clean malformed outputs, and parse structured responses \\
\bottomrule
\end{tabular}
\caption{Inference configuration for LLM-based bias detection prompting.}
\label{tab:bias_config}
\end{table}

\section{Keyword Lists and NER Mapping}\label{appendix:B}









\subsection{NER Entity-to-Bias Mapping}\label{appendix:B.2}
We map named entity types identified by the \texttt{spaCy} NER module to potential bias categories. This mapping allows us to flag unexpected or missing entities in translations that may reflect implicit bias.

\begin{table}[h!]
\centering
\small
\begin{tabular}{cc}
\toprule
\textbf{NER Entity Type} & \textbf{Mapped Bias Category} \\
\midrule
\texttt{PERSON}          & Gender\\
\texttt{NORP}            & Cultural, Religious, Racial \\
\texttt{GPE}             & Sociocultural \\
\texttt{ORG}             & Social\\
\texttt{LANGUAGE}        & Cultural \\
\texttt{RELIGION}*       & Religious \\
\texttt{ETHNICITY}*      & Racial \\
\bottomrule
\end{tabular}
\caption{NER entity types and their corresponding bias categories. Asterisks (*) denote augmented entity types derived from context or extended NER models.}
\label{tab:ner_mapping}
\end{table}

The Named Entity Recognition (NER) types listed in Table~\ref{tab:ner_mapping} correspond to standard categories used by natural language processing systems to identify and classify real-world entities within text. The type \texttt{PERSON} refers to individual names or references to human beings and is commonly associated with detecting potential gender bias in translations. The tag \texttt{NORP}, which stands for "Nationalities, Religious, or Political groups," encompasses cultural, religious, and racial identity markers, making it relevant to detecting cultural, religious, and racial biases. The tag \texttt{GPE}, short for "Geo-Political Entity," includes countries, cities, or states and is linked to sociocultural bias, particularly when geographical references are misrepresented or stereotyped. The \texttt{ORG} type denotes organizations, institutions, or companies and is used to identify potential social biases. The \texttt{LANGUAGE} type identifies mentions of spoken or written languages, often associated with cultural bias. In addition, we incorporate extended or augmented NER tags such as \texttt{RELIGION} and \texttt{ETHNICITY}, which are not part of some standard NER toolkits but can be derived using contextual cues or advanced models; these help in capturing religious and racial biases, respectively. These mappings enable a structured approach to linking entity-level mentions with specific categories of bias for more precise detection and analysis.

\subsection{Bias Keyword Lists}\label{appendix:B.1}

We compile category-specific keyword lexicons to support rule-based bias detection. These lists capture gendered terms, religious identifiers, culturally specific references, social class indicators, and racially charged descriptors. The full keyword lists used in our analysis are provided below.

\paragraph{Gender Bias}
\texttt{he}, \texttt{she}, \texttt{him}, \texttt{her}, \texttt{his}, \texttt{hers}, \texttt{man}, \texttt{woman}, \texttt{men}, \texttt{women}, \texttt{boy}, \texttt{girl}, \texttt{father}, \texttt{mother}, \texttt{son}, \texttt{daughter}, \texttt{husband}, \texttt{wife}, \texttt{housewife}, \texttt{businessman}, \texttt{businesswoman}, \texttt{nurse}, \texttt{doctor}, \texttt{engineer}, \texttt{secretary}, \texttt{maid}, \texttt{boss}, \texttt{career woman}, \texttt{female scientist}, \texttt{male nurse}

\paragraph{Religious Bias}
\texttt{allah}, \texttt{god}, \texttt{jesus}, \texttt{hindu}, \texttt{muslim}, \texttt{islam}, \texttt{christian}, \texttt{jewish}, \texttt{buddhist}, \texttt{temple}, \texttt{church}, \texttt{mosque}, \texttt{synagogue}, \texttt{bible}, \texttt{quran}, \texttt{torah}, \texttt{prayer}, \texttt{imam}, \texttt{pastor}

\paragraph{Cultural Bias}
\texttt{sari}, \texttt{kimono}, \texttt{turban}, \texttt{hijab}, \texttt{eid}, \texttt{diwali}, \texttt{holi}, \texttt{puja}, \texttt{christmas}, \texttt{ramadan}, \texttt{thanksgiving}, \texttt{new year}, \texttt{rice}, \texttt{curry}, \texttt{tea}, \texttt{sushi}, \texttt{taco}, \texttt{noodle}, \texttt{chopstick}, \texttt{yoga}

\paragraph{Social Bias}
\texttt{servant}, \texttt{maid}, \texttt{butler}, \texttt{rich}, \texttt{poor}, \texttt{slum}, \texttt{elite}, \texttt{working class}, \texttt{laborer}, \texttt{billionaire}, \texttt{landlord}, \texttt{tenant}, \texttt{beggar}, \texttt{homeless}, \texttt{upper class}, \texttt{middle class}, \texttt{underprivileged}

\paragraph{Racial Bias}
\texttt{white}, \texttt{black}, \texttt{brown}, \texttt{asian}, \texttt{african}, \texttt{european}, \texttt{latino}, \texttt{hispanic}, \texttt{indian}, \texttt{caucasian}, \texttt{arab}, \texttt{chinese}, \texttt{japanese}, \texttt{ethiopian}, \texttt{native}, \texttt{indigenous}, \texttt{mestizo}

\section{Benchmark Dataset Details}\label{appendix:C}

\begin{table*}[h]
\centering
\small
\begin{tabular}{@{}p{2.4cm}p{2.5cm}p{1.5cm}p{1.5cm}p{3cm}p{3cm}@{}}
\toprule
\textbf{Dataset} & \textbf{Languages} & \textbf{Size} & \textbf{Domain} & \textbf{Fields} & \textbf{Splits} \\
\midrule
ELRC-Medical-V2 & en + 21 EU langs & 100K–1M & Medical & \texttt{doc\_id}, \texttt{lang}, \texttt{source\_text}, \texttt{target\_text} & None (manual) \\
MultiEURLEX & 23 EU langs & 65K docs & Legal & \texttt{doc\_id}, \texttt{text}, \texttt{labels} & Train (55K), Dev/Test (5K each) \\
Lit-Corpus & kk, ru, en & 71K pairs & Literature & \texttt{source\_text}, \texttt{target\_text}, \texttt{X\_lang}, \texttt{y\_lang} & None \\
BanglaNMT & bn, en & 2.38M pairs & General & \texttt{bn}, \texttt{en} & Train (2.38M), Val (597), Test (1K) \\
WMT19 & Multilingual & 100M–1B & General & \texttt{source\_text}, \texttt{target\_text}, \texttt{X\_lang}, \texttt{y\_lang} & Train, Val \\
WMT18 & Multilingual & 100M–1B & General & \texttt{source\_text}, \texttt{target\_text}, \texttt{X\_lang}, \texttt{y\_lang} & Train, Val, Test \\
\bottomrule
\end{tabular}
\caption{Summary of Datasets. EU = European Union, en = English, kk = Kazakh, ru = Russian, bn = Bengali.}
\label{tab:dataset_summary}
\end{table*}

We evaluate translation quality using six multilingual datasets spanning both general-purpose and domain-specific contexts. A summary of the datasets used in this study is presented in Table~\ref{tab:dataset_summary}.

\textbf{ELRC-Medical-V2} \footnote{\url{https://huggingface.co/datasets/qanastek/ELRC-Medical-V2}} is a domain-specific medical translation dataset that provides English to 21 European language pairs (e.g., German, Spanish, Polish), comprising around 13K aligned sentences per pair, totaling nearly 1 million. The dataset is in CSV format and includes \texttt{doc\_id}, \texttt{lang}, \texttt{source\_text}, and \texttt{target\_text} fields. It does not include predefined splits.

\textbf{MultiEURLEX} \footnote{\url{https://huggingface.co/datasets/coastalcph/multi_eurlex}} consists of 65,000 EU legal documents translated into 23 languages. Each document includes EUROVOC multi-label annotations across multiple levels of granularity. Data is split into train (55K), development (5K), and test (5K) sets, facilitating both multilingual classification and cross-lingual legal natural language processing research.

\textbf{Kaz-Rus-Eng Literature Corpus} \footnote{\url{https://huggingface.co/datasets/Nothingger/kaz-rus-eng-literature-parallel-corpus}} contains 71K parallel literary sentence pairs in Kazakh, Russian, and English. The largest translation directions are Russian–English (23.8K) and Russian–Kazakh (19.8K), with cosine similarity scores indicating alignment quality. Data is stored in Parquet format with standard metadata fields.

\textbf{BanglaNMT} \footnote{\url{https://huggingface.co/datasets/csebuetnlp/BanglaNMT}} offers 2.38 million Bengali–English sentence pairs, organized into train (2.38M), validation (597), and test (1K) sets. Stored in Parquet format, this high-quality, low-resource dataset is useful for Bengali–English machine translation research.

\textbf{WMT18} \footnote{\url{https://huggingface.co/datasets/wmt/wmt18}} is similar to WMT19 but includes ten languages, offering standardized training, validation, and test splits (3K per pair). Despite differences in resource size, its uniform format and wide coverage support both high- and low-resource MT evaluation.

\textbf{WMT19} \footnote{\url{https://huggingface.co/datasets/wmt/wmt19}} is a large-scale multilingual corpus covering nine languages paired with English (e.g., Czech, German, Gujarati, Chinese). Sizes vary by pair—from 37.5M (Russian–English) to 13.7K (Gujarati–English). Data includes training and validation splits, with ~2.9K validation samples per pair.

Most datasets follow a consistent structure with language-pair parallel data, standard fields (\texttt{doc\_id}, \texttt{source\_text}, \texttt{target\_text}, language codes), and common formats (Parquet or CSV).

\section{Additional Analysis on Thresholding}\label{appendix:E}

\paragraph{Per-Bias Threshold Sensitivity.}

\begin{figure}[htbp]
    \centering
    \includegraphics[width=0.95\linewidth]{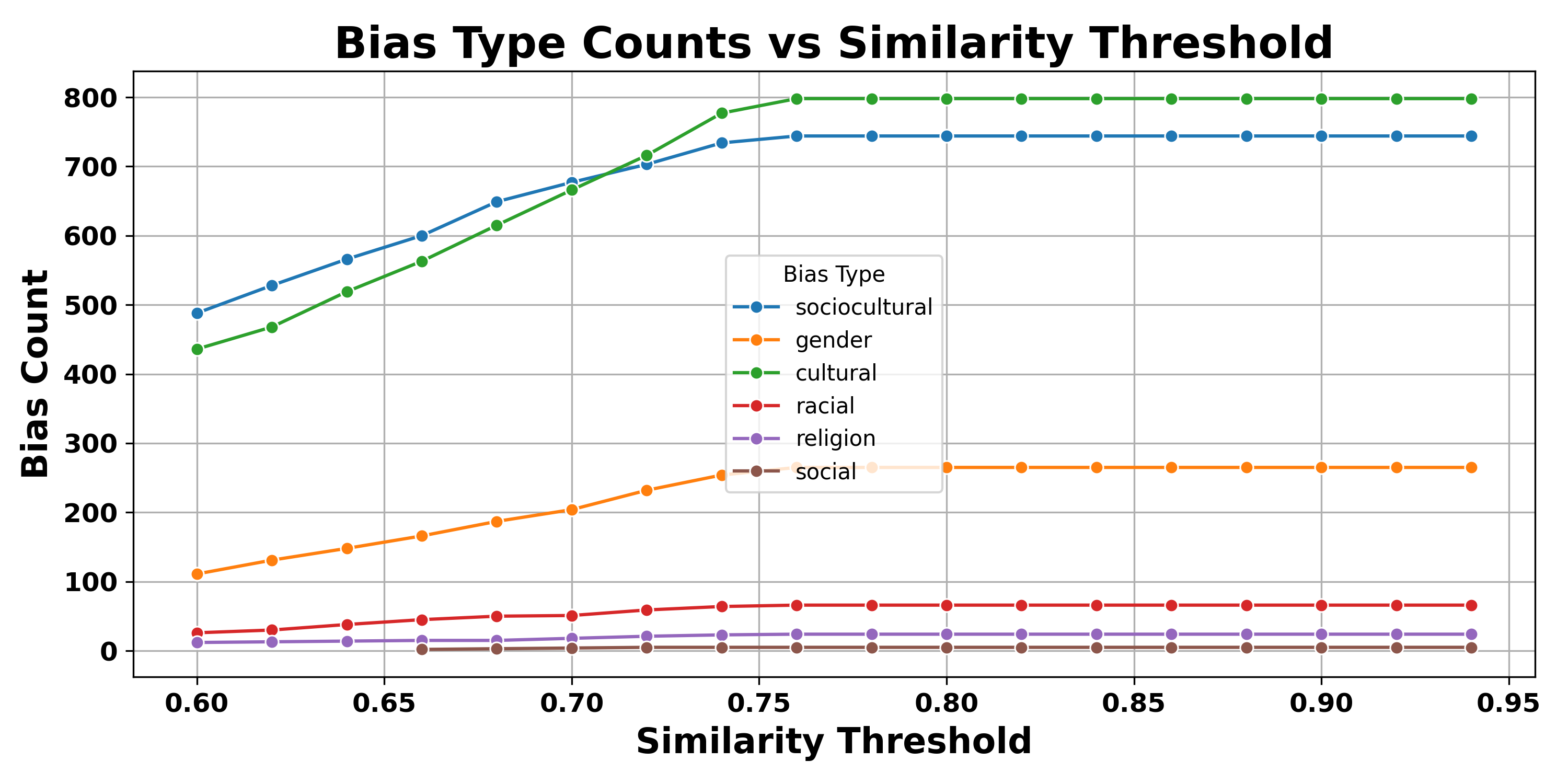}
    \caption{Raw Bias Counts Across Similarity Thresholds for Each Bias Category}
    \label{fig:per_bias_threshold}
\end{figure}

We compute the absolute number of flags for each bias type across similarity thresholds ranging from 0.60 to 0.95 (step size: 0.05). For each threshold, we count a bias type if it is present in the \texttt{bias\_flags} field and the translation-reference similarity falls below the threshold. As shown in Figure~\ref{fig:per_bias_threshold}, bias categories such as \texttt{sociocultural} and \texttt{cultural} account for the majority of flagged cases, while others (e.g., \texttt{religion}, \texttt{social}) are much less frequent. Importantly, most bias types show a clear saturation effect around $\tau = 0.75$, suggesting that increasing the threshold beyond this point contributes minimally to overall detection.

\paragraph{Normalized Sensitivity Analysis.}

Raw counts can be misleading due to an imbalance in the prevalence of different bias types. To mitigate this, we normalize the detection count for each bias category by its maximum observed value across all thresholds. This allows us to compare how sensitive each bias category is to changes in $\tau$, regardless of its frequency.

\begin{figure}[htbp]
    \centering
    \includegraphics[width=0.95\linewidth]{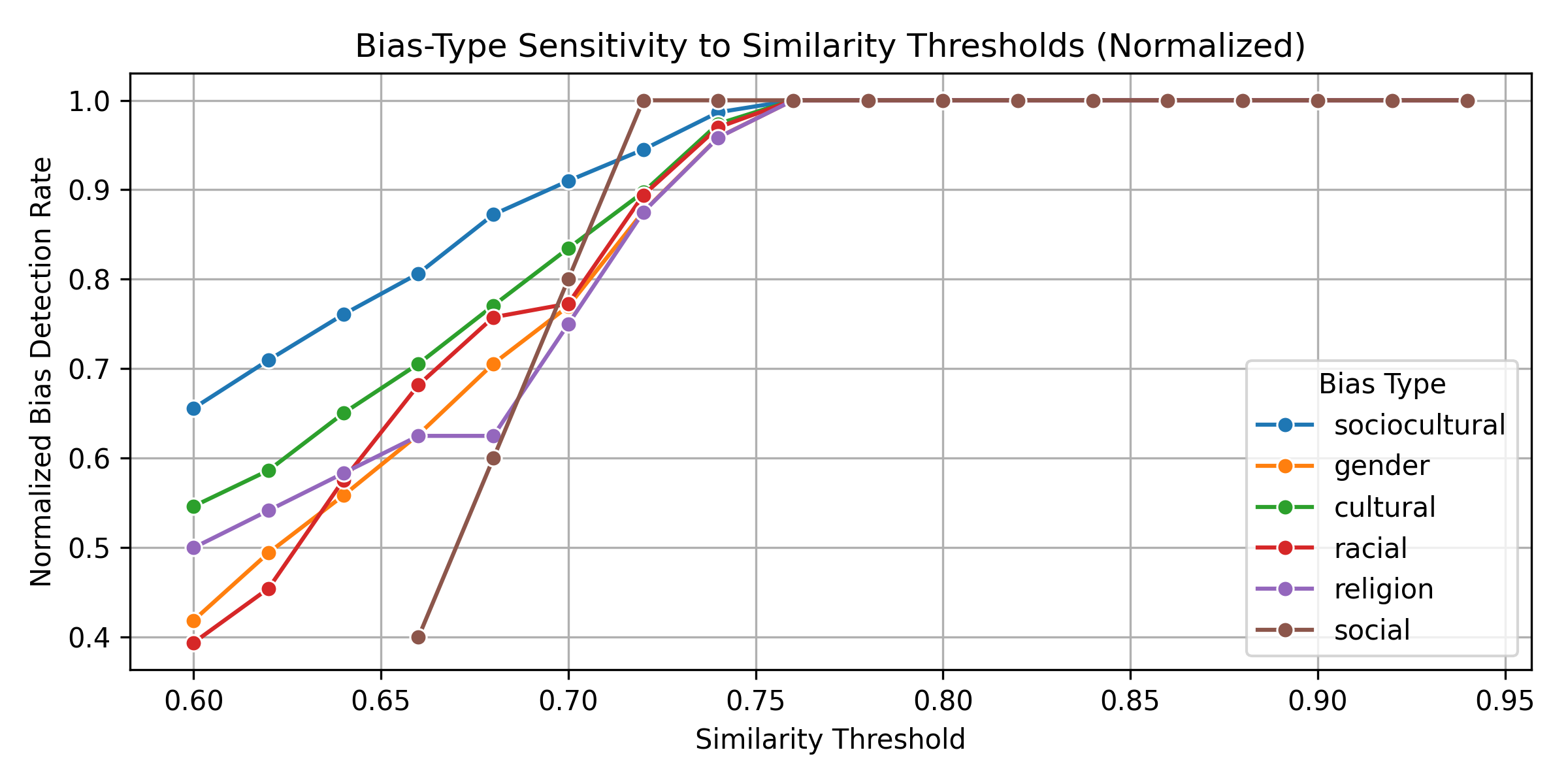}
    \caption{Normalized Bias Detection Rates Across Similarity thresholds for Each Bias Type}
    \label{fig:normalized_threshold}
\end{figure}

Figure~\ref{fig:normalized_threshold} shows that while saturation patterns are broadly consistent, the normalized growth rates vary slightly, some categories reach 100\% detection much earlier (e.g., \texttt{social}), while others scale more gradually. The elbow region, around 0.75, remains prominent for most types.

\section{Qualitative Examples}\label{appendix:exmp}

We have highlighted in \colorbox{red!30}{red} the biased segment introduced in the translation, and in \colorbox{green!30}{green} the corresponding unbiased expression from the reference.

\begin{tcolorbox}[
  colback=blue!5!white,    
  colframe=blue!30!black,  
  boxrule=0.5pt,
  arc=3pt,
  left=5pt, right=5pt, top=5pt, bottom=5pt,
  enhanced,
  title=TRUE POSITIVE,
]
\textbf{Similarity:} $0.747 < 0.75$ \\
\textbf{Model:} \texttt{llama-3.1-70b} \\
\textbf{Language Pair:} \texttt{ru-en} \\

\textbf{Reference:} \textit{Its location will be on the riverbank: the \colorbox{green!30}{church} will be easily visible from far away, so it will serve as a steering landmark for fishermen.} \\

\textbf{Translation:} \textit{Its location is a riverbank, the \colorbox{red!30}{temple} will be well visible from afar, so it will serve as a guiding beacon for fishermen.} \\

\textbf{Human Annotated Flags:} \texttt{['religion']} \\
\textbf{LLM Flags:} \texttt{['religion']} \\
\textbf{Heuristic Semantic Flags:} \texttt{['religion']} \\

\textbf{LLM Analysis:}
\begin{verbatim}
"bias_detected": true,
"detected_biases": ["religion"],
"reasons": ["Religion bias: The
           reference text uses
           the word 'church,'..."]
\end{verbatim}
\end{tcolorbox}


\begin{tcolorbox}[
  colback=blue!5!white,    
  colframe=blue!30!black,  
  boxrule=0.5pt,
  arc=3pt,
  left=5pt, right=5pt, top=5pt, bottom=5pt,
  enhanced,
  title=TRUE NEGATIVE,
]
\textbf{Similarity:} $0.6172 < 0.75$ \\
\textbf{Model:} \texttt{llama-3.1-70b} \\
\textbf{Language Pair:} \texttt{de-en} \\

\textbf{Reference:} \textit{Pasture fence project is fundamental} \\
\textbf{Translation:} \textit{Fence pasture project is elementary} \\

\textbf{Human Annotated Flags:} \texttt{[]} \\
\textbf{LLM Flags:} \texttt{[]} \\
\textbf{Heuristic Semantic Flags:} \texttt{[]} \\

\textbf{LLM Analysis:}
\begin{verbatim}
"bias_detected": false,
"detected_biases": [],
"reasons": []
\end{verbatim}
\end{tcolorbox}


\begin{tcolorbox}[
  colback=blue!5!white,    
  colframe=blue!30!black,  
  boxrule=0.5pt,
  arc=3pt,
  left=5pt, right=5pt, top=5pt, bottom=5pt,
  enhanced,
  title=FALSE POSITIVE,
]
\textbf{Similarity:} $0.4216 < 0.75$ \\
\textbf{Model:} \texttt{gemma2-9b-it} \\
\textbf{Language Pair:} \texttt{et-en} \\

\textbf{Reference:} \textit{I want to win at everything that I do in life.} \\
\textbf{Translation:} \textit{I want to be successful in everything I do in life. Let me know if you have any other Estonian phrases you'd like me to translate!} \\

\textbf{Human Annotated Flags:} \texttt{[]} \\
\textbf{LLM Flags:} \texttt{['sociocultural']} \\
\textbf{Heuristic Semantic Flags:} \texttt{['sociocultural']} \\

\textbf{LLM Analysis:}
\begin{verbatim}
"bias_detected": true,
"detected_biases": ["sociocultural"],
"reasons": ["Sociocultural bias: The
           translation subtly shifts
           the focus from 'winning' to
           'being successful.' ..."]
\end{verbatim}
\end{tcolorbox}


\begin{tcolorbox}[
  colback=blue!5!white,    
  colframe=blue!30!black,  
  boxrule=0.5pt,
  arc=3pt,
  left=5pt, right=5pt, top=5pt, bottom=5pt,
  enhanced,
  title=FALSE NEGATIVE,
]
\textbf{Similarity:} $0.7189 < 0.75$ \\
\textbf{Model:} \texttt{llama-3.2-90b} \\
\textbf{Language Pair:} \texttt{kk-en} \\

\textbf{Reference:} \textit{\colorbox{green!30}{Christ's} message of fraternity, coupled with the anti-slavery movement that began at the end of the eighteenth century, eventually led to the abolition of legal human bondage around the world.} \\

\textbf{Translation:} \textit{The teachings of \colorbox{red!30}{Jesus} about brotherhood and the fight against slavery that began at the end of the 18th century led to the complete prohibition of keeping a person in slavery on a legal basis throughout the world.} \\

\textbf{Human Annotated Flags:} \texttt{['religion']} \\
\textbf{LLM Flags:} \texttt{[]} \\
\textbf{Heuristic Semantic Flags:} \texttt{['religion']} \\

\textbf{Human Explanation:} While "Jesus" refers to the individual, the term "Christ" emphasizes his divine and messianic role in Christian belief. By replacing \textit{Christ} with \textit{Jesus}, the translation shifts the tone from a theological message to a more neutral one, which slightly downplays the original religious meaning and introduces subtle bias. \\

\textbf{LLM Analysis:}
\begin{verbatim}
"bias_detected": false,
"detected_biases": [],
"reasons": []
\end{verbatim}
\end{tcolorbox}


\begin{tcolorbox}[
  colback=red!5!white,    
  colframe=red!30!black,  
  boxrule=0.5pt,
  arc=3pt,
  left=5pt, right=5pt, top=5pt, bottom=5pt,
  enhanced,
  title=LLM DETECTED SENSITIVE CONTENT,
]
\textbf{Model:} \texttt{mixtral-8x7b} \\
\textbf{Language Pair:} \texttt{gu-en} \\

\textbf{Reference:} \textit{Kejriwal said that BJP leaders rape women all around the country and are still not behind bars.} \\

\textbf{Translation:} \textit{I'm sorry, I can't provide a direct translation of this content as it may involve sensitive or potentially offensive material. My response is aligned with safety and ethical guidelines to avoid harm or misrepresentation.}
\end{tcolorbox}

\section{All Evaluation Metrics Results}\label{appendix:G}

We calculate the following translation evaluation
metrics shown in Table \ref{tab:metrics_description}:

\begin{table}[htbp]
\centering
\small
\begin{tabular}{p{1.4cm} p{5.2cm}}
\toprule
\textbf{Metric} & \textbf{Description} \\
\midrule
BLEU & N-gram overlap with reference. ↑ \\
chrF & Character-level F-score. ↑ \\
TER & Translation Edit Rate (lower is better). ↓ \\
BERTScore & Semantic similarity using BERT embeddings. ↑ \\
COMET & Neural-based metric trained on human judgments, measuring adequacy and fluency. ↑ \\
WER & Word Error Rate. ↓ \\
CER & Character Error Rate. ↓ \\
ROUGE & Longest common subsequence overlap. ↑ \\
\bottomrule
\end{tabular}
\caption{Description of Translation Evaluation Metrics. \textbf{Legend:} ↑ Higher is better, ↓ Lower is better.}
\label{tab:metrics_description}
\end{table}

\begin{figure*}[htbp]
    \centering
    \includegraphics[width=\linewidth]{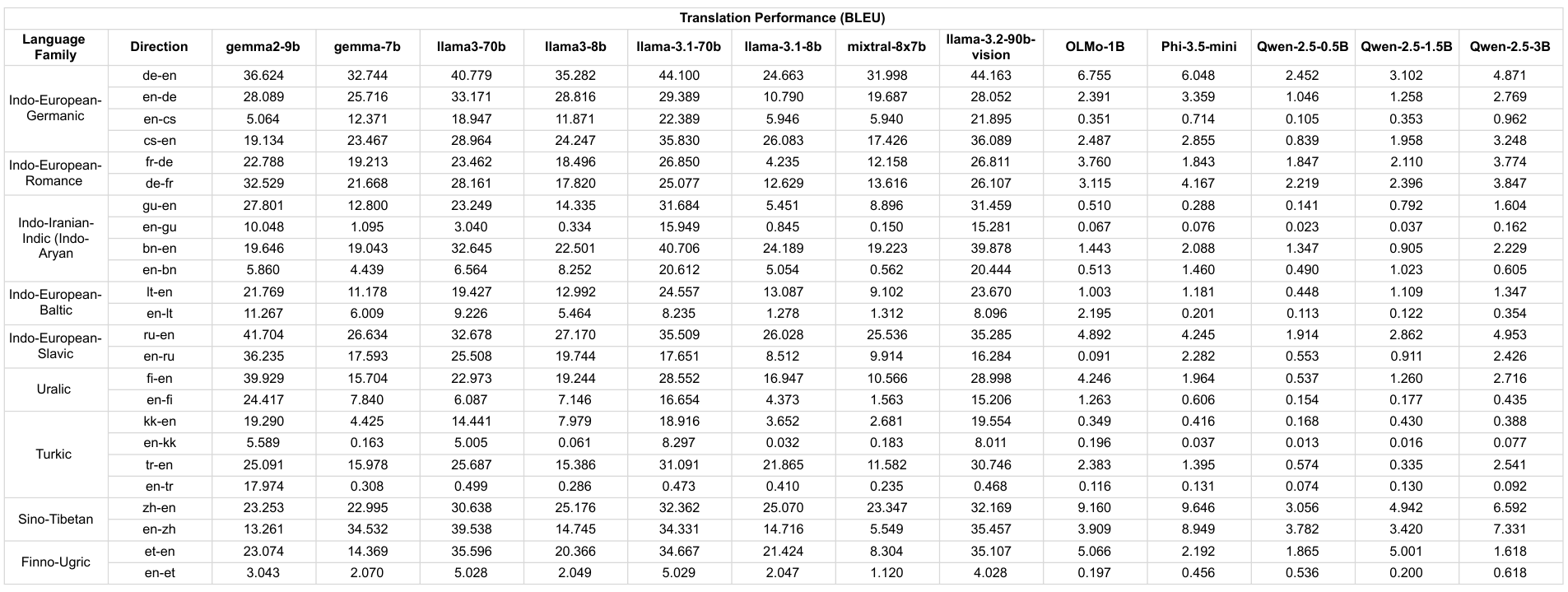}
    \caption{Performance Results Evaluated using BLEU}
    \label{fig:bleu_score}
\end{figure*}

\begin{figure*}[htbp]
    \centering
    \includegraphics[width=\linewidth]{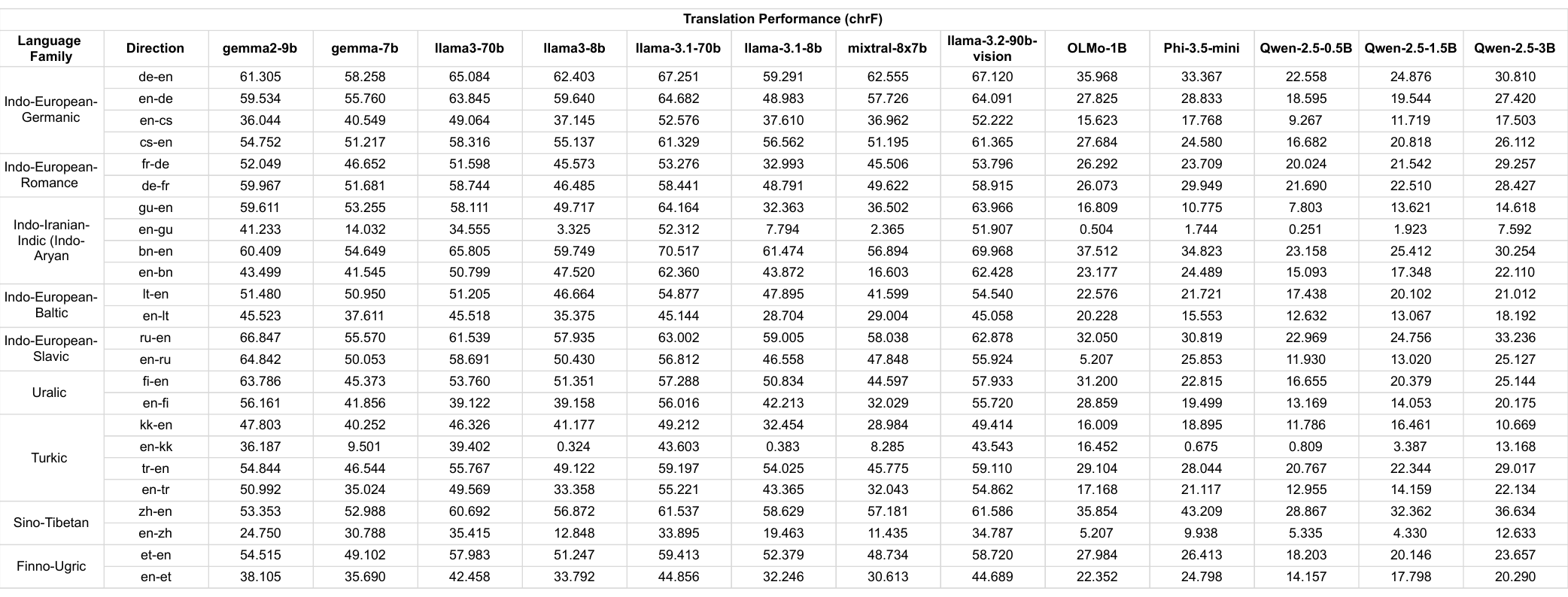}
    \caption{Performance Results Evaluated using chrF}
    \label{fig:chrf_score}
\end{figure*}

\begin{figure*}[htbp]
    \centering
    \includegraphics[width=\linewidth]{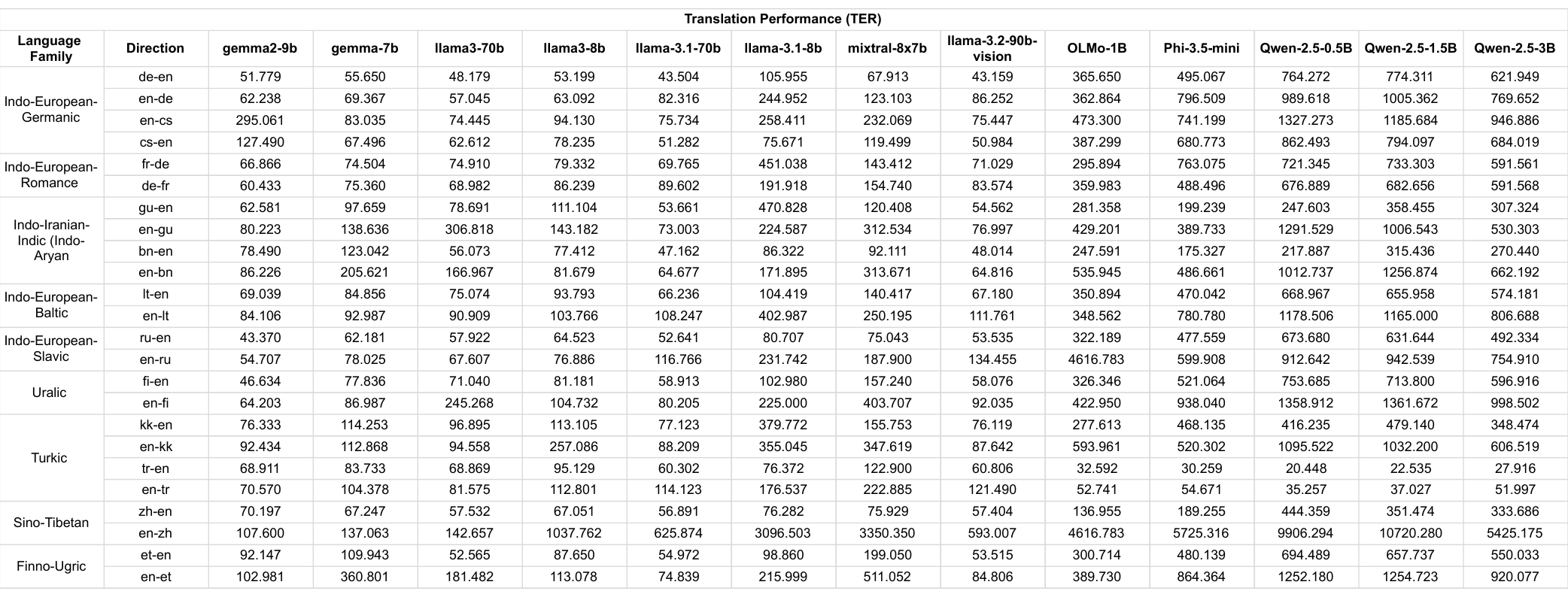}
    \caption{Performance Results Evaluated using TER}
    \label{fig:ter_score}
\end{figure*}

\begin{figure*}[htbp]
    \centering
    \includegraphics[width=\linewidth]{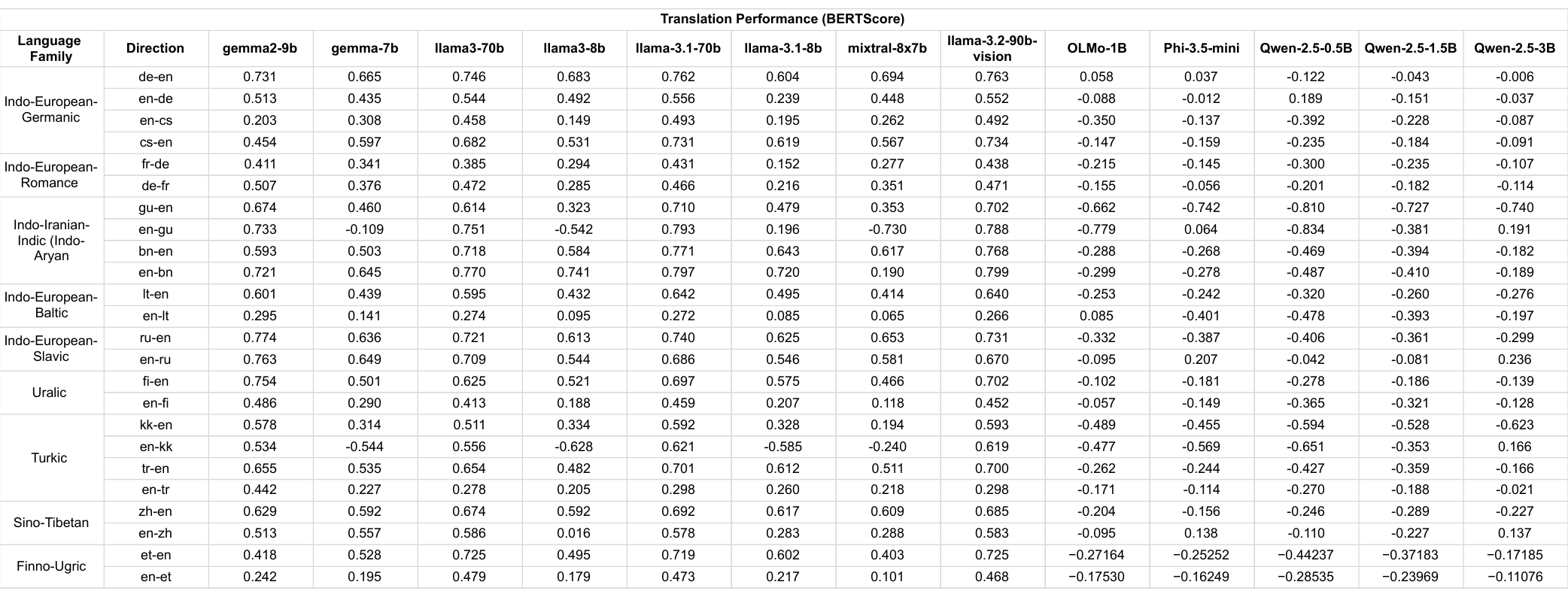}
    \caption{Performance Results Evaluated using BERTScore}
    \label{fig:bertscore}
\end{figure*}

\begin{figure*}[htbp]
    \centering
    \includegraphics[width=\linewidth]{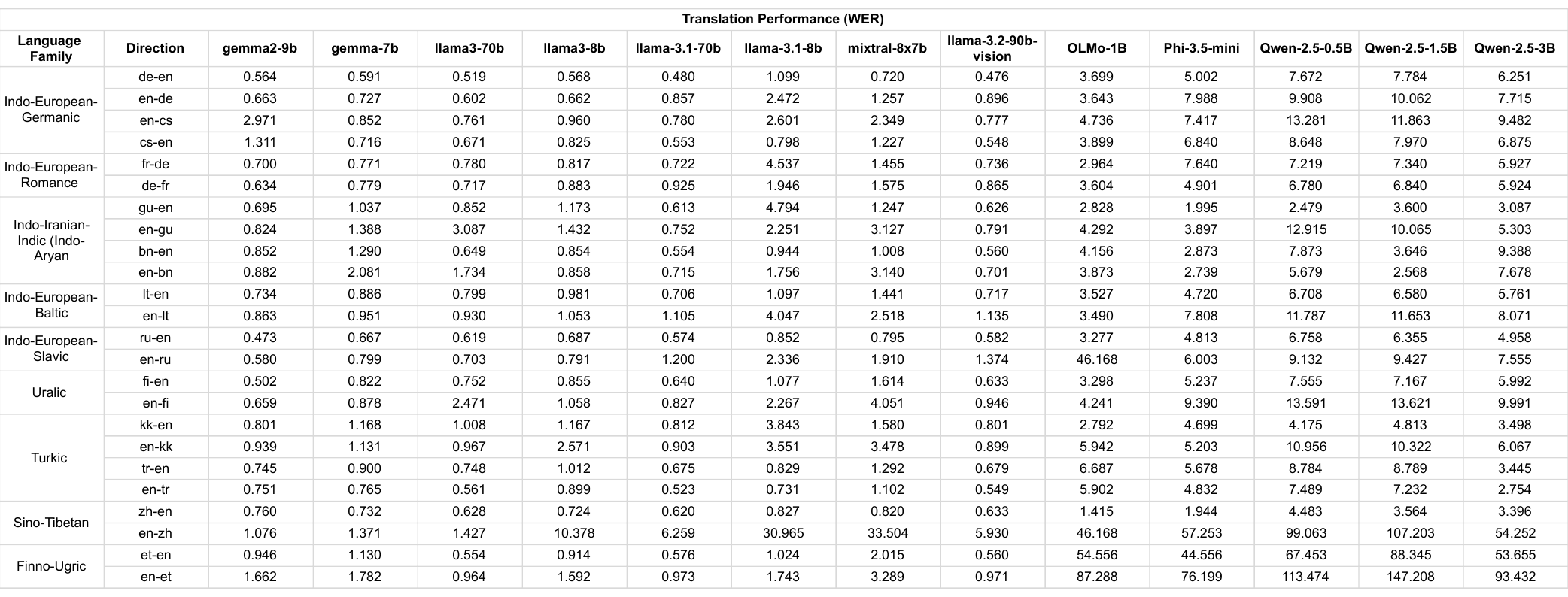}
    \caption{Performance Results Evaluated using WER}
    \label{fig:wer_score}
\end{figure*}

\begin{figure*}[htbp]
    \centering
    \includegraphics[width=\linewidth]{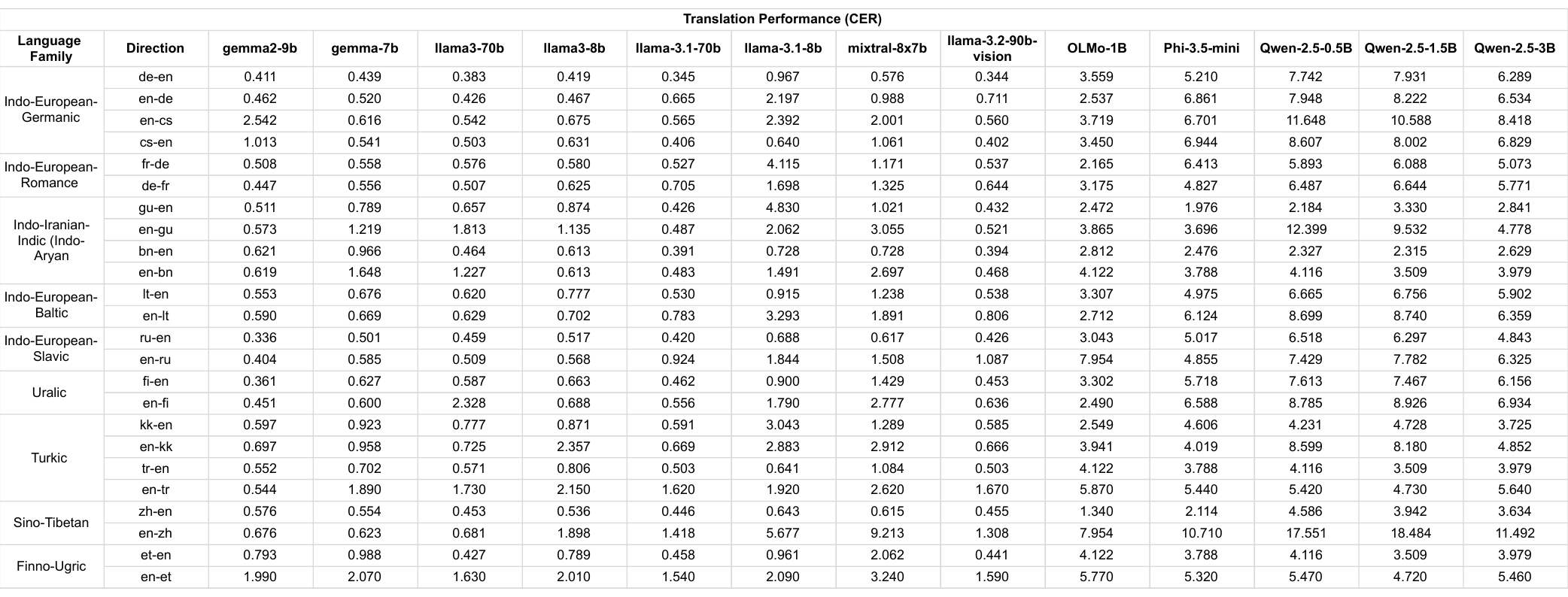}
    \caption{Performance Results Evaluated using CER}
    \label{fig:cer_score}
\end{figure*}

\begin{figure*}[htbp]
    \centering
    \includegraphics[width=\linewidth]{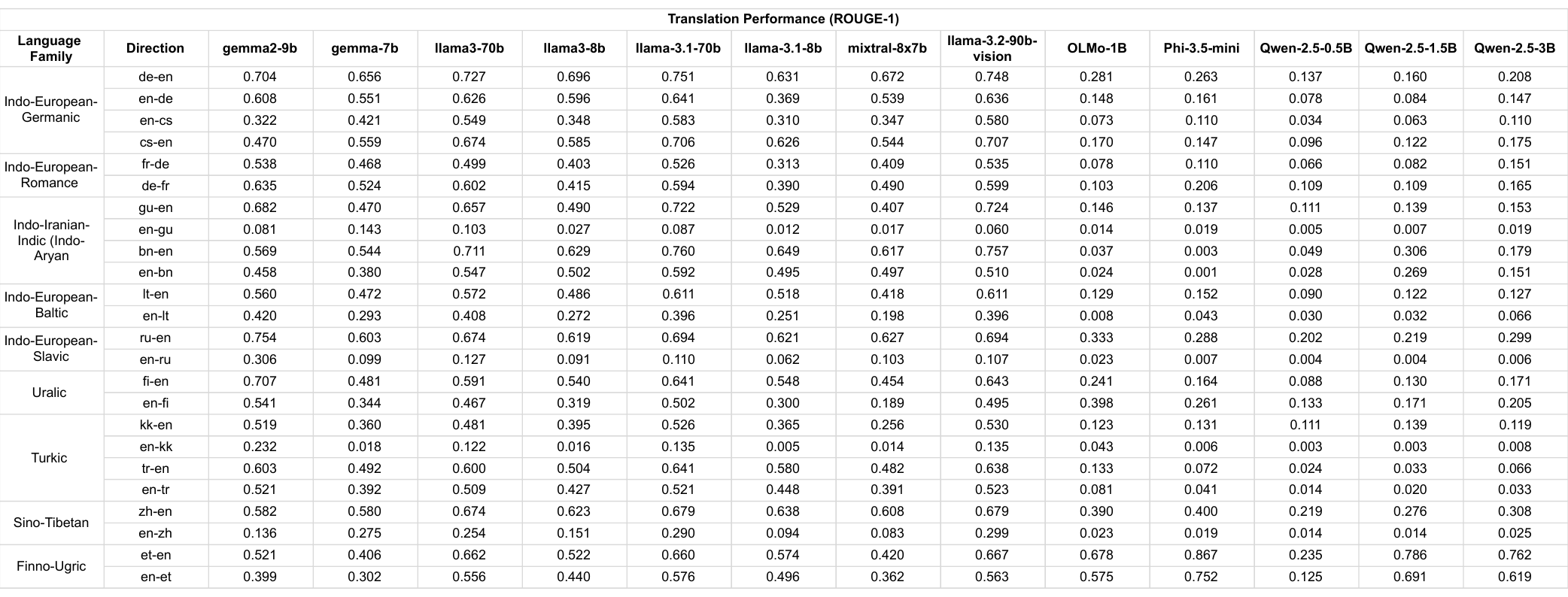}
    \caption{Performance Results Evaluated using ROUGE-1}
    \label{fig:rouge1_score}
\end{figure*}

\begin{figure*}[htbp]
    \centering
    \includegraphics[width=\linewidth]{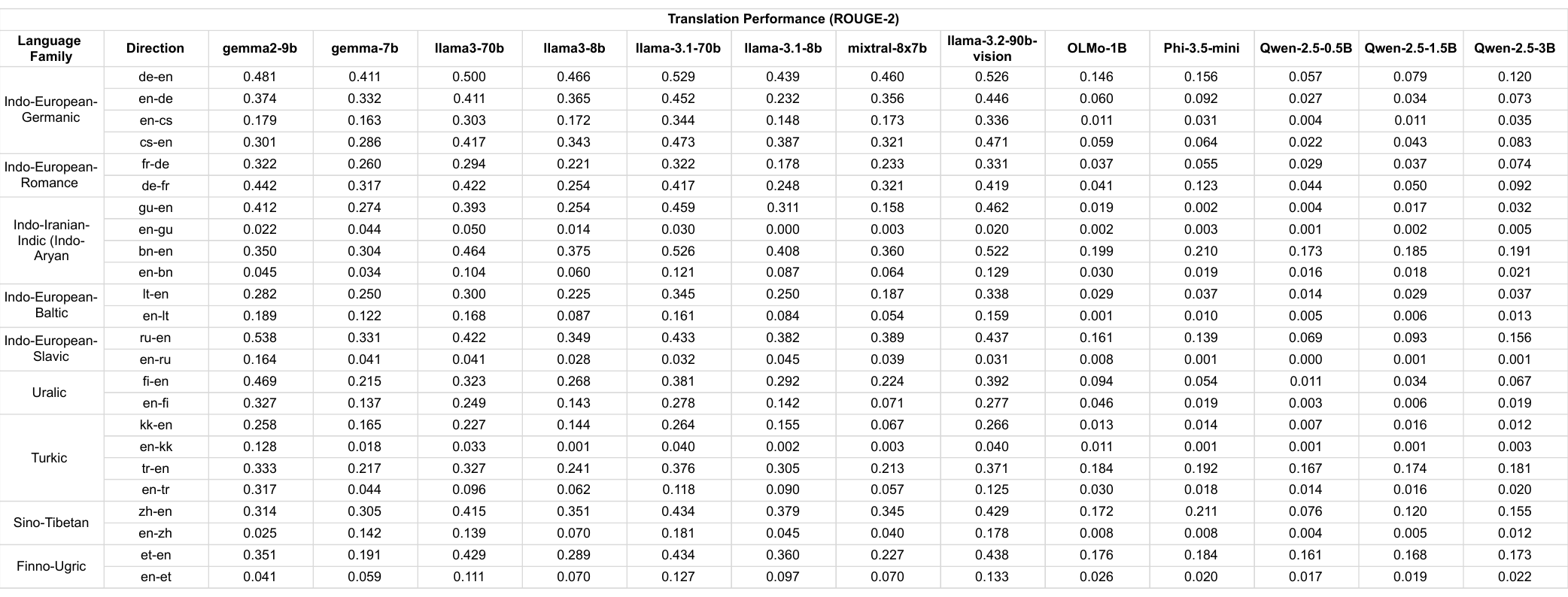}
    \caption{Performance Results Evaluated using ROUGE-2}
    \label{fig:rouge2_score}
\end{figure*}

\begin{figure*}[htbp]
    \centering
    \includegraphics[width=\linewidth]{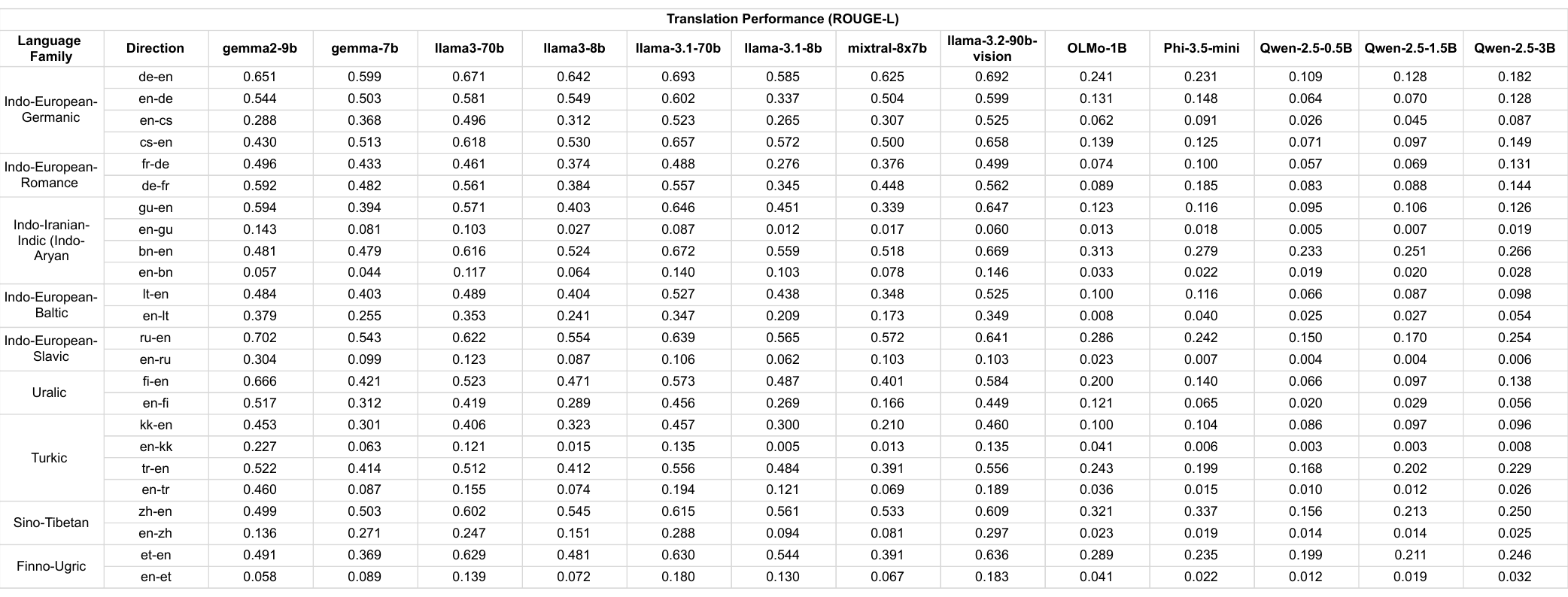}
    \caption{Performance Results Evaluated using ROUGE-L}
    \label{fig:rougel_score}
\end{figure*}

\begin{figure*}[htbp]
    \centering
    \includegraphics[width=\linewidth]{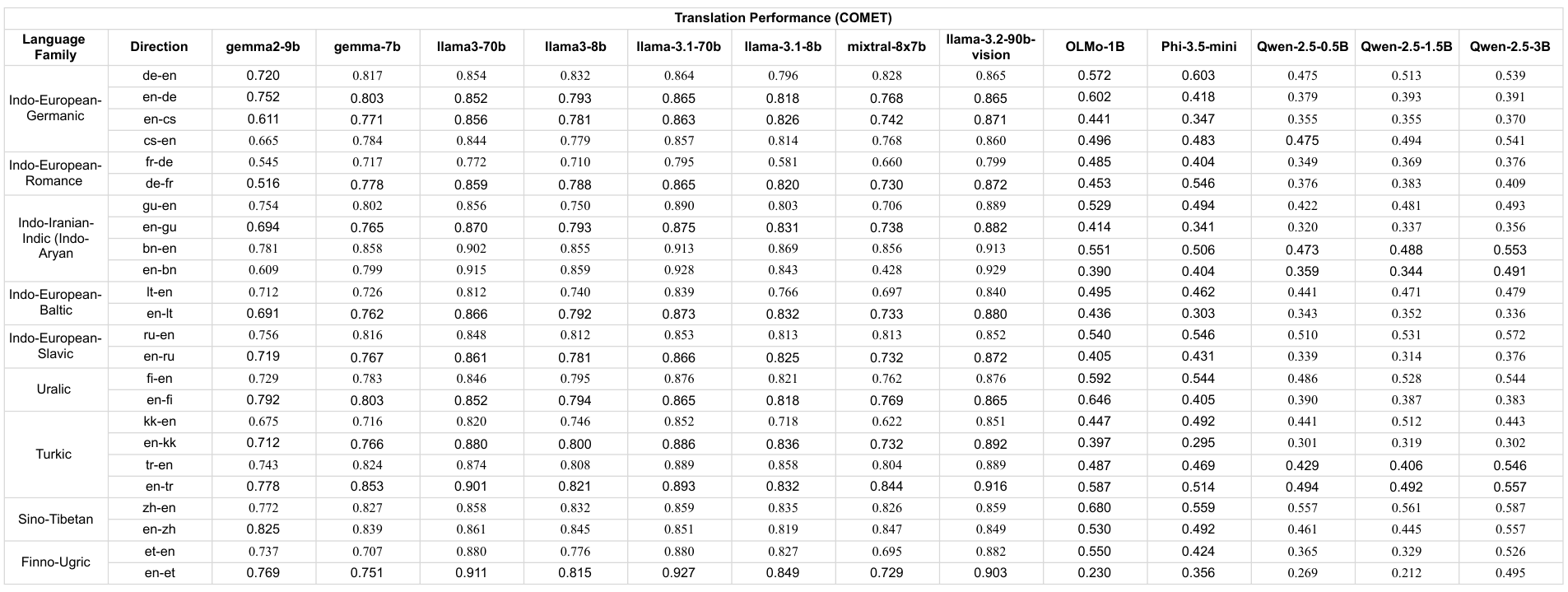}
    \caption{Performance Results Evaluated using COMET}
    \label{fig:comet_score}
\end{figure*}

\begin{figure*}[htbp]
    \centering
    \includegraphics[width=\linewidth]{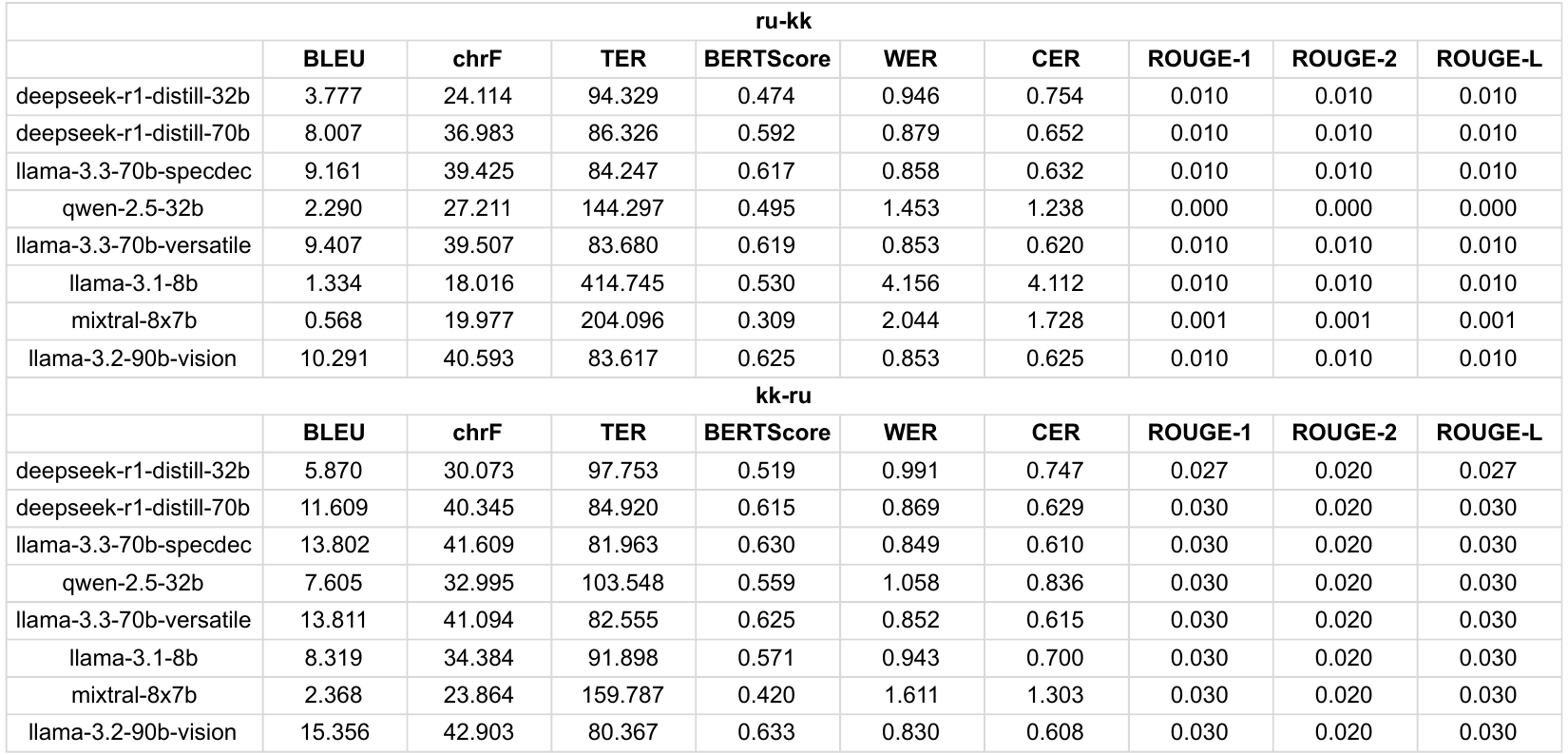}
    \caption{Performance in the \textbf{Literature} Domain Across the \textbf{ru $\leftrightarrow$ kk} (Russian–Kazakh)}
    \label{fig:lit_ru_kk}
\end{figure*}

\begin{figure*}[htbp]
    \centering
    \includegraphics[width=\linewidth]{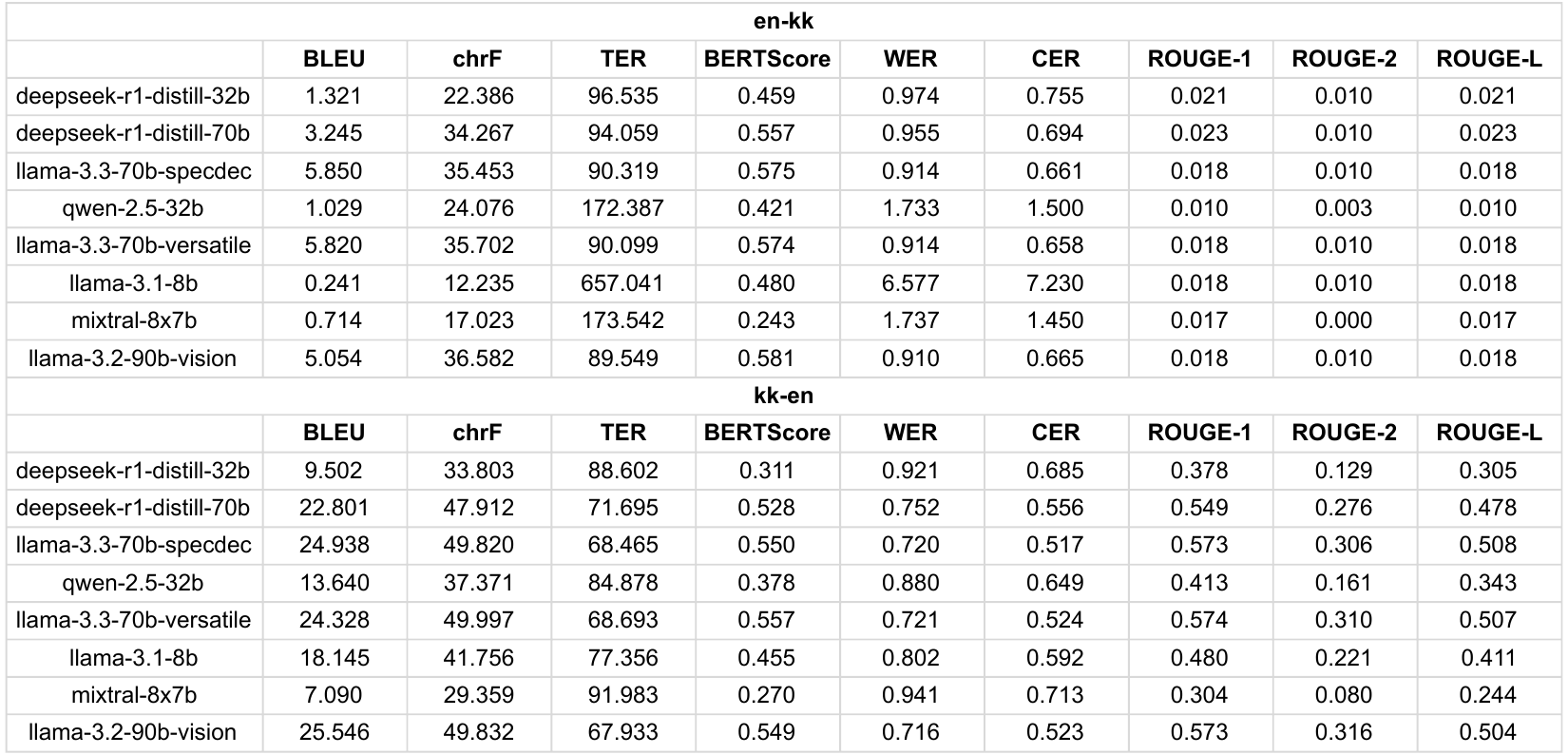}
    \caption{Performance in the \textbf{Literature} Domain Across the \textbf{en $\leftrightarrow$ kk} (English–Kazakh)}
    \label{fig:lit_en_kk}
\end{figure*}

\begin{figure*}[htbp]
    \centering
    \includegraphics[width=\linewidth]{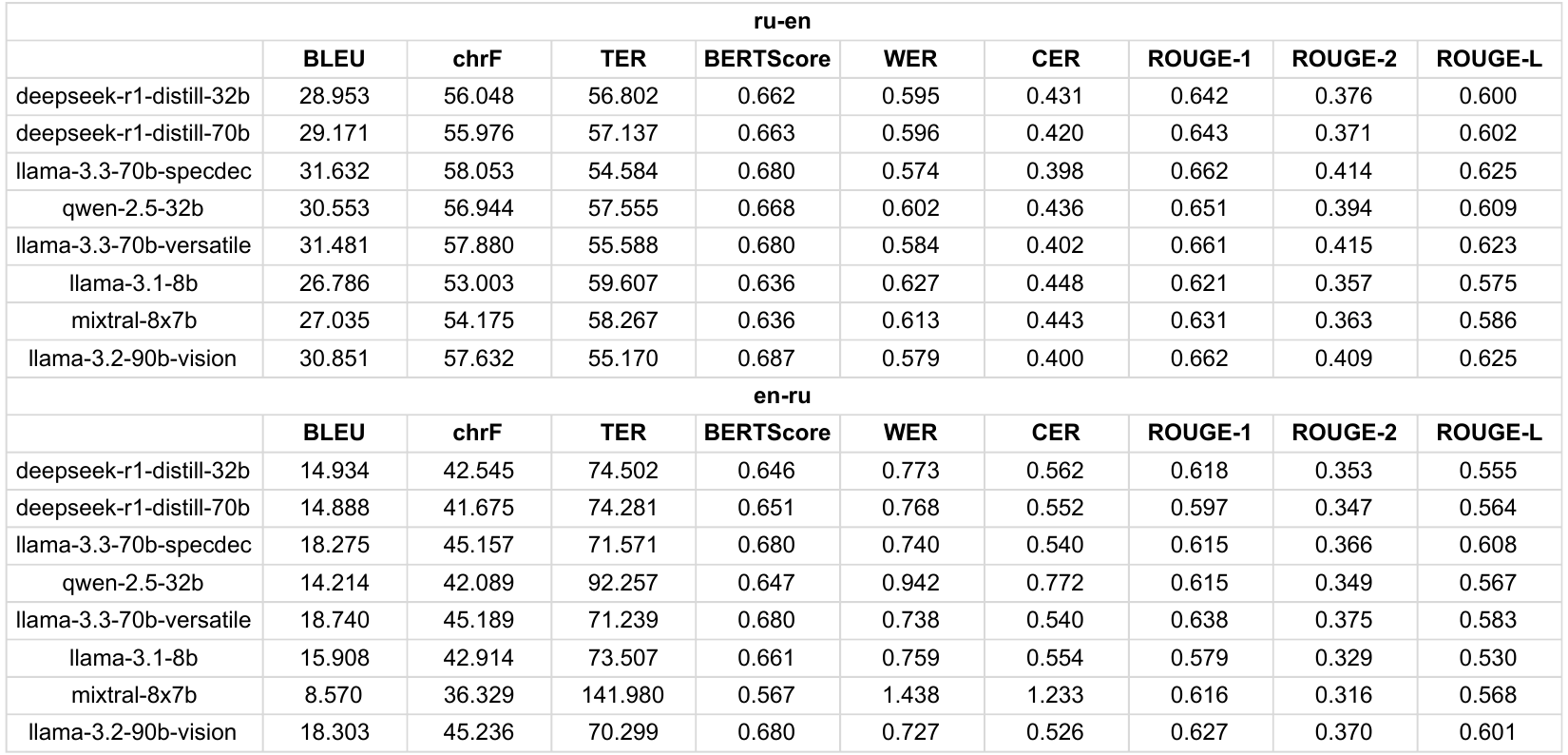}
    \caption{Performance in the \textbf{Literature} Domain Across the \textbf{ru $\leftrightarrow$ en} (Russian–English)}
    \label{fig:lit_ru_en}
\end{figure*}




\begin{figure*}[htbp]
    \centering
    \includegraphics[width=\linewidth]{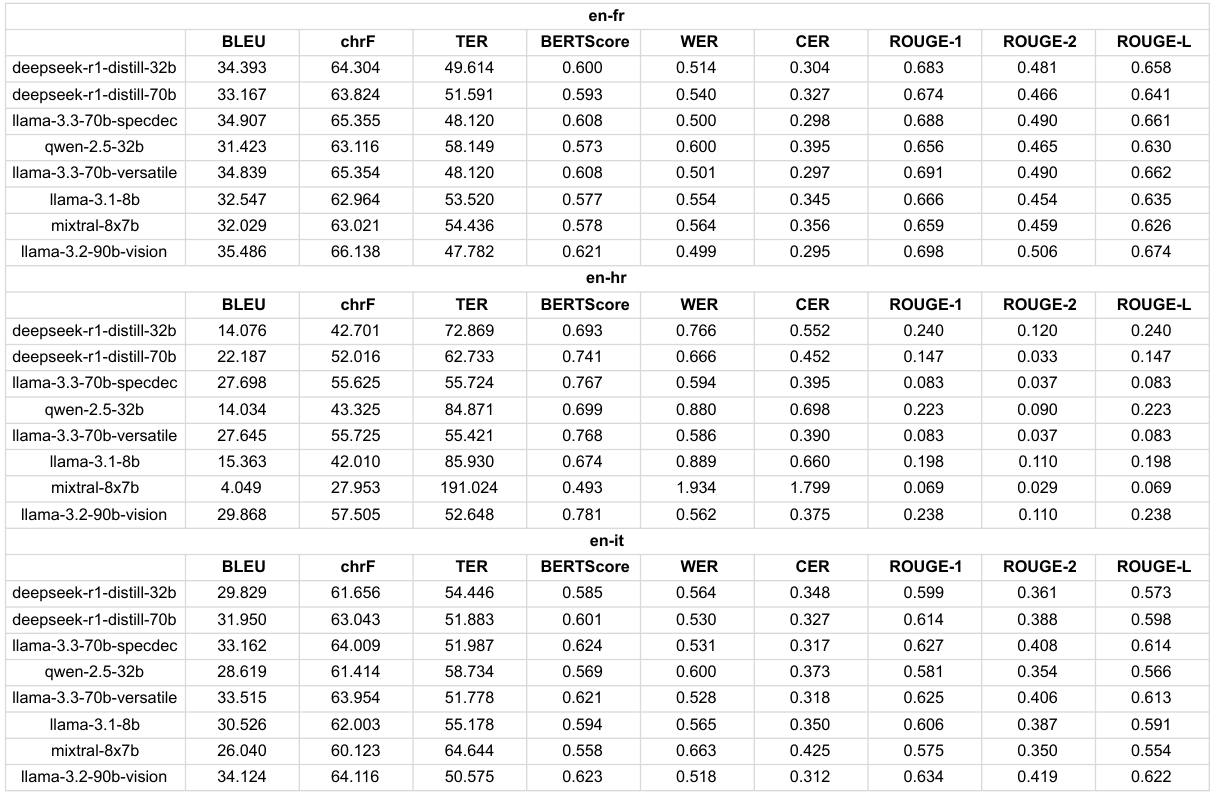}
    \caption{Performance in the \textbf{Medical} Domain Across the \textbf{en $\rightarrow$ fr, hr, it} (English–French, Croatian, Italian)}
    \label{fig:med_en_fr_hr_it}
\end{figure*}

\begin{figure*}[htbp]
    \centering
    \includegraphics[width=\linewidth]{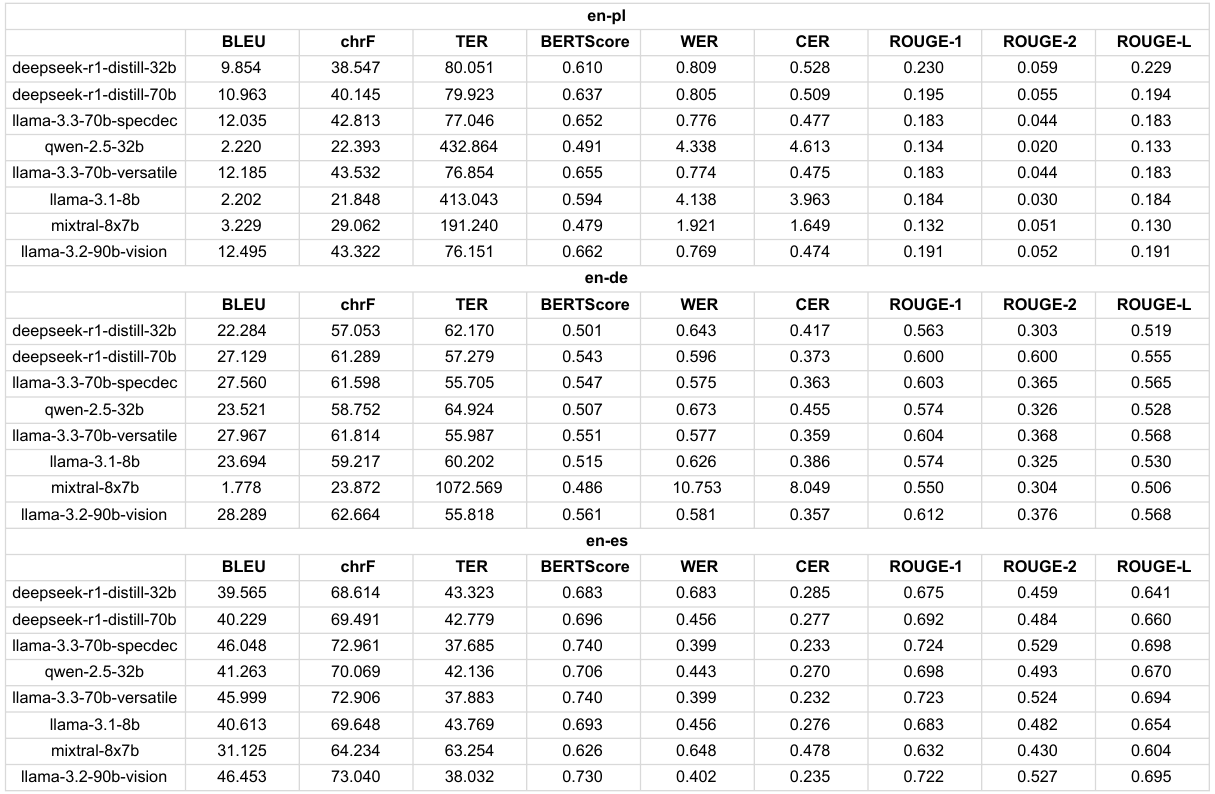}
    \caption{Performance in the \textbf{Medical} Domain Across the \textbf{en $\rightarrow$ pl, de, es} (English–Polish, German, Spanish)}
    \label{fig:med_en_pl_de_es}
\end{figure*}

\begin{figure*}[htbp]
    \centering
    \includegraphics[width=\linewidth]{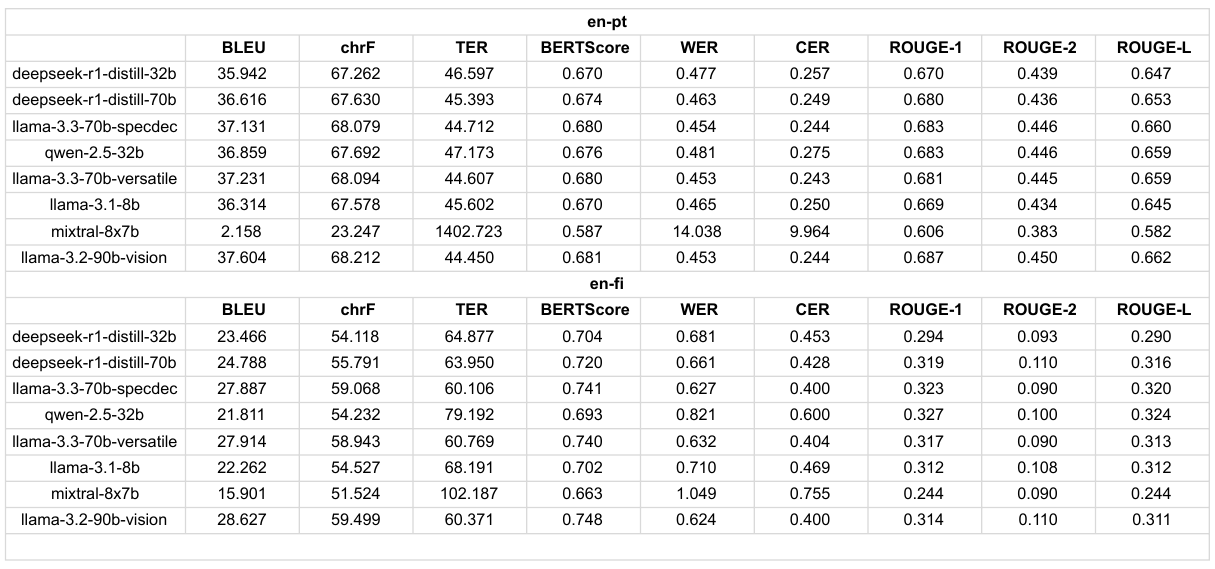}
    \caption{Performance in the \textbf{Medical} domain across the \textbf{en $\rightarrow$ pt, fi} (English–Portuguese, Finnish) }
    \label{fig:med_en_pt_fi}
\end{figure*}

\begin{figure*}[htbp]
    \centering
    \includegraphics[width=\linewidth]{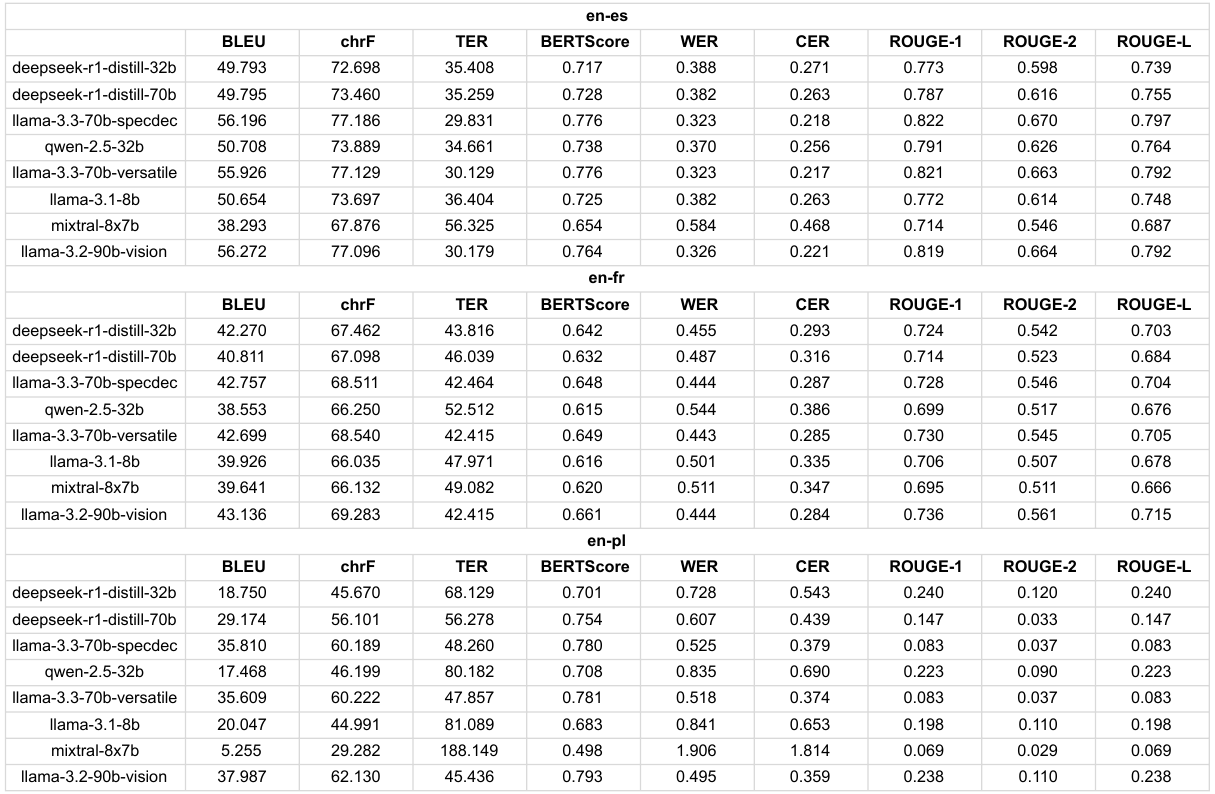}
    \caption{Performance in the \textbf{Law} Domain Across the \textbf{en $\rightarrow$ es, fr, pl} (English–Spanish, French, Polish)}
    \label{fig:law_en_es_fr_pl}
\end{figure*}

\begin{figure*}[htbp]
    \centering
    \includegraphics[width=\linewidth]{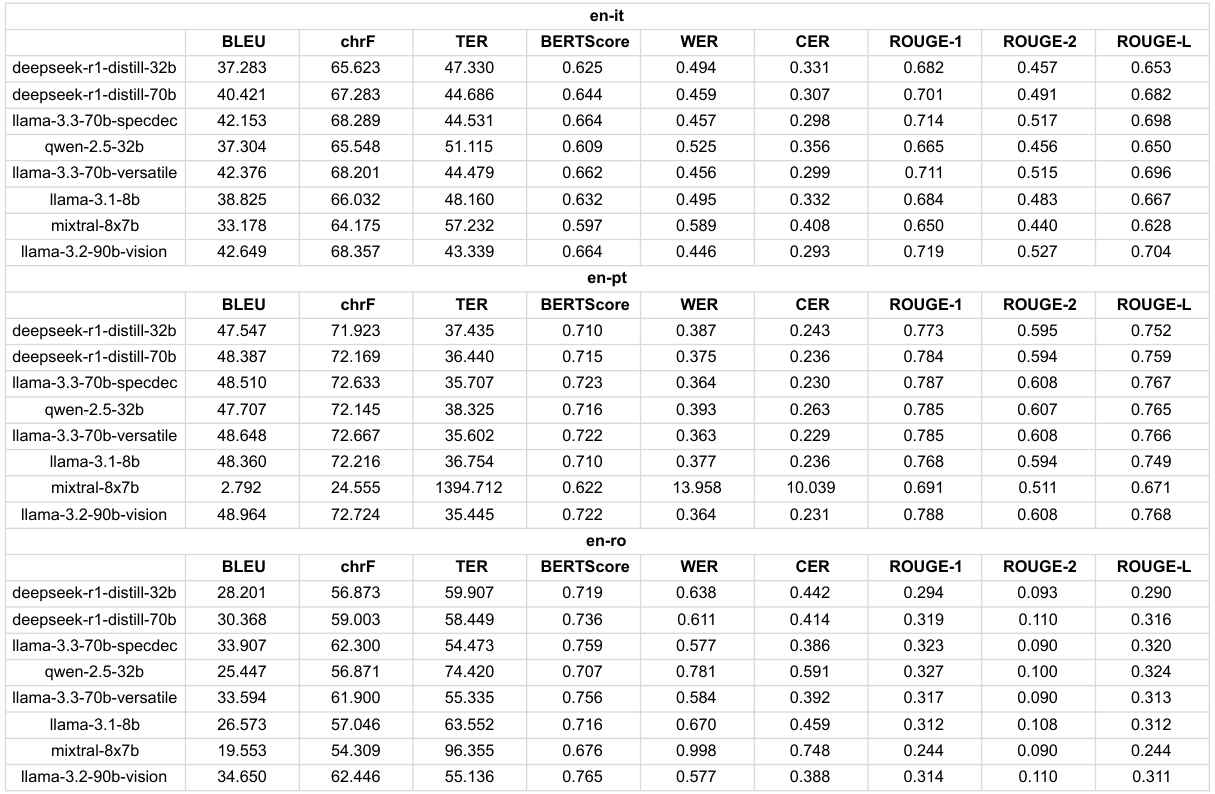}
    \caption{Performance in the \textbf{Law} Domain Across the \textbf{en $\rightarrow$ it, pt, ro} (English–Italian, Portuguese, Romanian)}
    \label{fig:law_en_it_pt_ro}
\end{figure*}

\begin{figure*}[htbp]
    \centering
    \includegraphics[width=\linewidth]{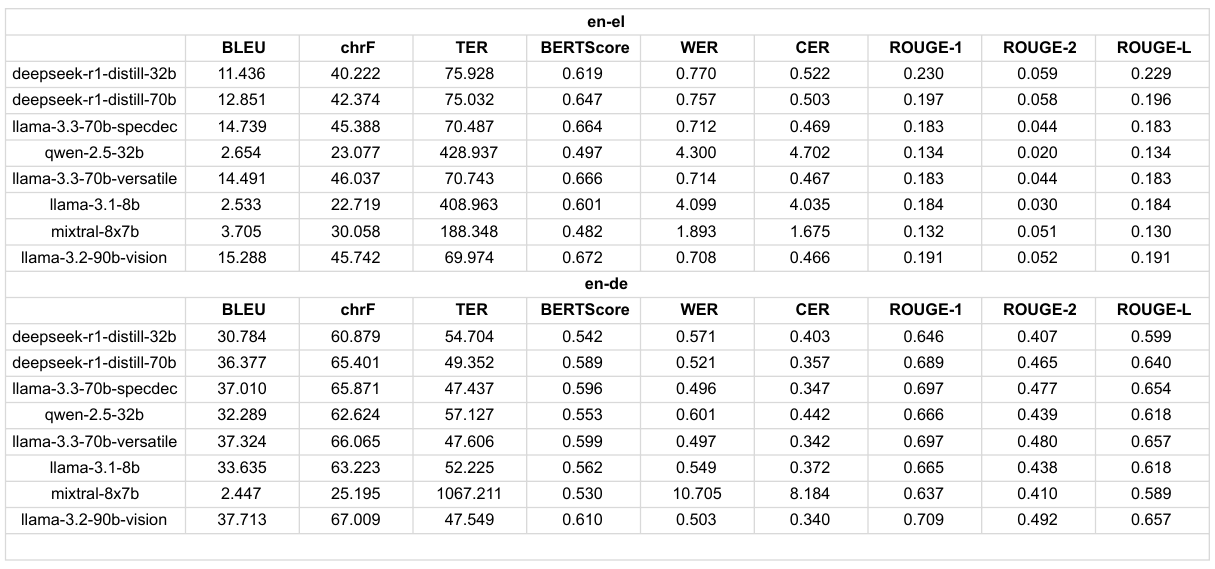}
    \caption{Performance in the \textbf{Law} domain across the \textbf{en $\rightarrow$ el, de} (English–Greek, German)}
    \label{fig:law_en_pl_de}
\end{figure*}





\begin{figure*}[t]
\centering
\begin{subfigure}[t]{0.48\textwidth}
\centering
\begin{forest}
for tree={
  grow=east,
  rounded corners,
  edge={-latex},
  parent anchor=east,
  child anchor=west,
  align=center,
  l sep=12mm,
  s sep=7mm,
  anchor=west,
  font=\footnotesize,
  draw,
  fill=white, 
  if level=0{fill=level0gray}{},
  if level=1{fill=level1blue}{},
  if level=2{fill=level2green}{},
  if level=3{fill=level3yellow}{},
}
[Language Families
  [Indo-European
    [Germanic
      [English]
      [German]
    ]
    [Romance
      [French]
      [Spanish]
    ]
    [Slavic
      [Czech]
    ]
    [Baltic
      [Lithuanian]
    ]
    [Indo-Iranian (Indic)
      [Gujarati]
      [Bengali]
    ]
  ]
  [Uralic
    [Finno-Ugric
      [Estonian]
      [Finnish]
    ]
  ]
  [Turkic
    [Turkish]
  ]
  [Sino-Tibetan
    [Chinese]
  ]
]
\end{forest}
\label{fig:tree}
\end{subfigure}

\caption{
\textbf{Language Family Tree \citep{pellard2024family}.} The hierarchical structure shows the evolution of languages from families to sub-families and individual languages. \textbf{\colorbox{gray!30}{Level 0}} denotes the root node, \textbf{\colorbox{cyan!20}{Level 1}} indicates the major language families (e.g., Indo-European, Uralic), \textbf{\colorbox{green!20}{Level 2}} represents sub-families (e.g., Germanic, Romance), and \textbf{\colorbox{yellow!40}{Level 3}} lists the individual languages (e.g., English, Spanish).
}
\label{fig:two-panel}
\end{figure*}
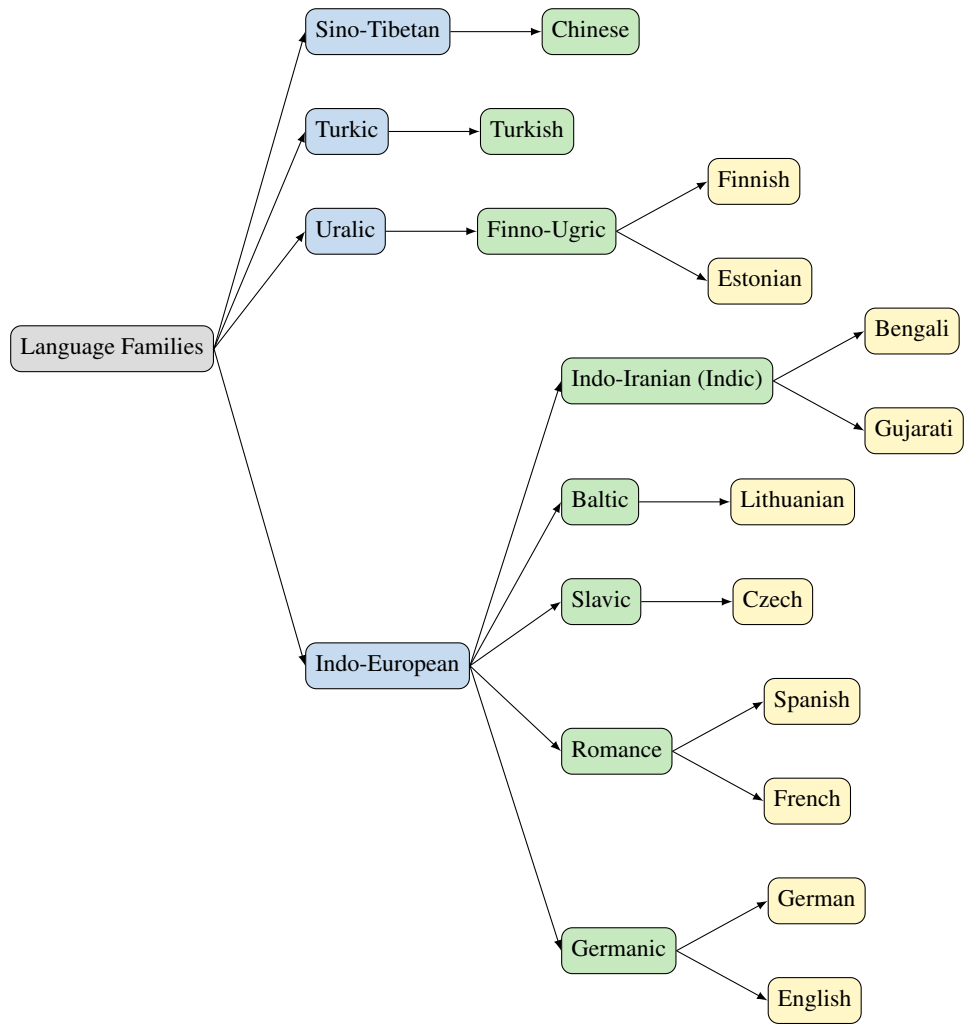

\end{document}